\documentclass[preprint,12pt]{elsarticle}

\usepackage{amsmath,amsfonts}
\usepackage{algorithmic}
\usepackage{array}
\usepackage[tight,footnotesize]{subfigure}
\usepackage{textcomp}
\usepackage{stfloats}
\usepackage{url}
\usepackage{verbatim}
\usepackage{graphicx}
\def\BibTeX{{\rm B\kern-.05em{\sc i\kern-.025em b}\kern-.08em
    T\kern-.1667em\lower.7ex\hbox{E}\kern-.125emX}}
\usepackage{balance}
\usepackage{booktabs}
\usepackage{caption}
\usepackage{makecell}
\usepackage{adjustbox}
\usepackage{wrapfig}
\usepackage[linesnumbered,ruled]{algorithm2e}
\usepackage{amssymb}
\usepackage{lineno}
\usepackage{float}
\usepackage{algorithmic}
\usepackage[tight,footnotesize]{subfigure}
\usepackage{booktabs}
\usepackage{caption}
\usepackage{makecell}
\usepackage{adjustbox}
\usepackage{wrapfig}
\usepackage{xcolor}
\usepackage[linesnumbered,ruled]{algorithm2e}

\journal{Journal Name}

\begin{document}
\sloppy
\setlength{\parskip}{0pt}

\begin{frontmatter}

\title{A Practical Guide for Designing, Developing, and Deploying Production-Grade Agentic AI Workflows}



\author[label1]{Eranga Bandara}
\ead{cmedawer@odu.edu}

\author[label1]{Ross Gore}
\ead{rgore@odu.edu}

\author[label1]{Peter Foytik}
\ead{pfoytik@odu.edu}

\author[label1]{Sachin Shetty}
\ead{sshetty@odu.edu}

\author[label1]{Ravi Mukkamala}
\ead{mukka@odu.edu}

\author[label2]{Abdul Rahman}
\ead{abdulrahman@deloitte.com}

\author[label3]{Xueping Liang}
\ead{xuliang@fiu.edu}

\author[label1]{Safdar H. Bouk}
\ead{sbouk@odu.edu}

\author[label8]{Amin Hass}
\ead{aminhass@analetiq.com}

\author[label7]{Sachini Rajapakse}
\ead{sachini.rajapakse@iciclelabs.ai}

\author[label4]{Ng Wee Keong}
\ead{awkng@ntu.edu.sg}

\author[label5]{Kasun De Zoysa}
\ead{kasun@ucsc.cmb.ac.lk}

\author[label9]{Aruna Withanage}
\ead{aruna@effectz.ai}
\author[label9]{Nilaan Loganathan}
\ead{nilaan@effectz.ai}

\address[label1]{Old Dominion University, Norfolk, VA, USA}
\address[label2]{Deloitte \& Touche LLP, USA}
\address[label3]{Florida International University, USA}
\address[label4]{Nanyang Technological University, Singapore}
\address[label5]{University of Colombo, Sri Lanka}
\address[label7]{IcicleLabs.AI}
\address[label8]{AnaletIQ, VA, USA}
\address[label9]{Effectz.AI}

\begin{abstract}

Agentic AI marks a major shift in how autonomous systems reason, plan, and execute multi-step tasks. Unlike traditional single model prompting, agentic workflows integrate multiple specialized agents with different Large Language Models(LLMs), tool-augmented capabilities, orchestration logic, and external system interactions to form dynamic pipelines capable of autonomous decision-making and action. As adoption accelerates across industry and research, organizations face a central challenge: how to design, engineer, and operate production-grade agentic AI workflows that are reliable, observable, maintainable, and aligned with safety and governance requirements. This paper provides a practical, end-to-end guide for designing, developing, and deploying production-quality agentic AI systems. We introduce a structured engineering lifecycle encompassing workflow decomposition, multi-agent design patterns, Model Context Protocol(MCP), and tool integration, deterministic orchestration, Responsible-AI considerations, and environment-aware deployment strategies. We then present nine core best practices for engineering production-grade agentic AI workflows, including tool-first design over MCP, pure-function invocation, single-tool and single-responsibility agents, externalized prompt management, Responsible-AI-aligned model-consortium design, clean separation between workflow logic and MCP servers, containerized deployment for scalable operations, and adherence to the Keep it Simple, Stupid (KISS) principle to maintain simplicity and robustness. To demonstrate these principles in practice, we present a comprehensive case study: a multimodal news-analysis and media-generation workflow. By combining architectural guidance, operational patterns, and practical implementation insights, this paper offers a foundational reference to build robust, extensible, and production-ready agentic AI workflows.

\end{abstract}


\end{frontmatter}

\section{Introduction}

\begin{figure}[H]
\centering{}
\includegraphics[width=5.3in]{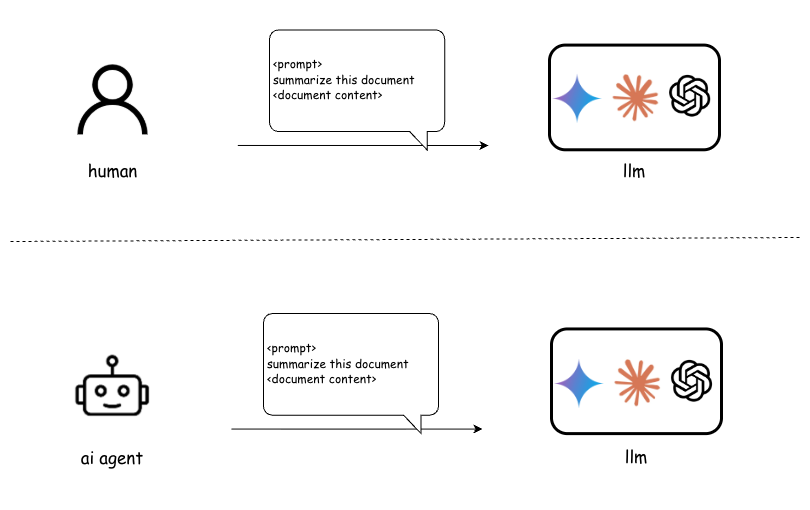}
\vspace{-0.1in}
\DeclareGraphicsExtensions.
\caption{Human–LLM interaction versus autonomous AI agent–LLM interaction.}
\label{ai-agent}
\end{figure}

The rapid advancement of Large Language Models (LLMs)~\cite{llm, gpt-llm}, Vision–Language Models (VLMs)~\cite{vision-language-model, pixtral, qwen2}, and tool-augmented reasoning has laid the foundation for a new paradigm in automation: agentic AI~\cite{agentic-ai, agentsway}. Traditional LLM interactions follow a simple pattern in which a human provides a prompt and the model generates a response (as illustrated in the top half of Figure~\ref{ai-agent}). In contrast, an AI agent can perform this same interaction autonomously: it can construct prompts, call models, interpret responses, and perform follow-up actions without direct human intervention (as illustrated in the bottom half of Figure~\ref{ai-agent}). In essence, AI agents are software programs that use LLMs—together with tools, APIs, and external context—to execute tasks automatically. When multiple such agents collaborate—each with a specialized role such as searching, filtering, scraping, reasoning, validating, or publishing—they form agentic AI workflows~\cite{agentsway}. These workflows enable systems that can reason, plan, take actions, monitor outcomes, retry intelligently, and iteratively refine their behavior. Modern agentic workflows integrate LLMs with external tools, structured memory, search functions, databases, Model Context Protocol (MCP) servers, cloud services, and API-driven environments~\cite{deep-stride}. Rather than relying on a single monolithic prompt, the system delegates responsibilities across specialized agents to ensure modularity, determinism, and maintainability. This evolution has produced dynamic extensible pipelines capable of solving high-value real-world automation problems—from content generation and news analytics, to regulatory compliance, knowledge extraction, reasoning pipelines, and multimodal media synthesis~\cite{proof-of-tbi}. By incorporating heterogeneous models (e.g., OpenAI, Gemini, Llama, Anthropic), deterministic tool calls, and environment-aware orchestration, agentic AI provides a flexible and powerful architectural approach for building scalable and production-grade AI systems.

However, building production-grade agentic AI workflows remains challenging. While prototypes are easy to build with simple scripts or notebooks, scaling them into reliable, governed, and observable systems introduces multiple engineering complexities. In particular, design challenges include issues such as ways to decompose workflows into agents, making choices between tool calls and MCP actions~\cite{mcp1, mcc}, ways to design deterministic orchestration, and methods to avoid implicit behaviors that lead to LLM drift or unpredictable execution paths~\cite{agentic-ai-workflow-patterns}. The implementation challenges include managing multi-agent communication, handling tool schemas, maintaining prompt modularity, integrating heterogeneous model families, and enforcing responsible AI principles while ensuring output consistency. The operational challenges include running workflows reliably in production environments; managing concurrency, failures, retries, logging, and cost efficiency; securing tool access; monitoring agent traces; and ensuring reproducibility across model updates. Finally, the deployment challenges include containerizing complex multi-model systems, integrating with Kubernetes, managing versioning, exposing services through APIs and MCP servers, and supporting continuous delivery of agent updates and prompt modifications~\cite{prompt-engineering, agentic-ai-challenges}.

As organizations move from experimentation to operational deployment, these challenges frequently result in brittle systems, inconsistent outputs, and high maintenance overhead. Without a disciplined engineering approach, agentic workflows can easily grow into opaque, unbounded, and error-prone pipelines that are difficult to debug, scale, or govern~\cite{agentic-ai-workflow-patterns}. This paper addresses these gaps by presenting a practical, engineering-focused guide for designing, developing, and deploying production-grade agentic AI workflows. Drawing from real-world implementations and large-scale deployments, we outline a structured methodology for workflow decomposition, agent specialization, tool integration, safety mechanisms, orchestration strategies, and environment-aware deployment. We complement these principles with nine actionable best practices aimed at improving determinism, reliability, observability, and maintainability. Finally, we demonstrate these principles in action through a fully implemented case study—a multimodal news-analysis and media-generation workflow that autonomously scrapes web content, filters topics, generates podcast scripts through an agent consortium~\cite{nurolense}, consolidates reasoning output, and produces audio and video artifacts before publishing them to GitHub. This example showcases how heterogeneous agents can be composed into a unified pipeline and deployed reliably in production. The following are our main contributions of this research.

\begin{enumerate}
    \item \textbf{A generalized engineering framework for production-grade agentic AI workflows.}  
    We introduce a structured methodology for designing, developing, and deploying agentic systems using multi-agent orchestration, tool integration, and deterministic execution patterns suitable for real-world automation.

    \item \textbf{A curated set of nine best practices for reliable and responsible–AI–enabled workflow design.}
    These practices encompass tool-over-MCP design, pure-function invocation, single-tool and single-responsibility agents, externalized prompt management, consortium-based reasoning for Responsible AI, separation of workflow logic and MCP servers, containerized deployment for scalability, maintainable workflow architecture, and adherence to the KISS principle for simplicity and robustness.

    \item \textbf{A full implementation of a multimodal, multi-agent news-to-media workflow.}  
    We present a complete case study demonstrating how the best practices are applied to a complex real-world workflow that performs feed discovery, topic filtering, content extraction, multi-LLM script generation, reasoning-based consolidation, audio/video synthesis, and automated GitHub publishing.

    \item \textbf{An extensible blueprint for organizations adopting agentic AI in production.}  
    By integrating containerization, Kubernetes orchestration, MCP accessibility, and Responsible-AI mechanisms (bias mitigation, reasoning audits, deterministic operations), the proposed system offers a robust template adaptable across domains such as compliance automation, media generation, analytics, and enterprise RPA.
\end{enumerate}

The remainder of the paper is organized as follows. Section 2 presents the motivating use case, showcasing an end-to-end agentic AI workflow for multimodal news analysis, content synthesis, and media generation. Section 3 builds upon this use case by introducing a curated set of best practices for designing, developing, and deploying production-grade agentic AI workflows, with emphasis on architectural choices, orchestration patterns, tooling strategies, and Responsible-AI–aligned design principles. Section 4 details the implementation of the proposed workflow—including agent design, tool/function integration, reasoning-based consolidation, multimodal media generation, and deployment in a containerized production environment. Section 5 evaluates the system, assessing the performance and accuracy of individual agents and the effectiveness of the reasoning-based consolidation across models. Finally, Section 6 concludes the paper by summarizing key insights and outlining directions for future research, standardization, and broader adoption of robust and trustworthy agentic AI systems.

\section{Use case: Podcast-Generation Agentic AI Workflow}

To illustrate the practical value and architectural complexity of production-grade agentic AI systems, this paper uses a real-world workflow for automated podcast generation from live news sources. The workflow integrates heterogeneous LLM agents, web-scraping tools, multimodal generation models, and a GitHub publishing pipeline into a fully autonomous end-to-end system. Figure 2 shows the complete architecture.

\begin{figure}[H]
\centering{}
\includegraphics[width=5.5in]{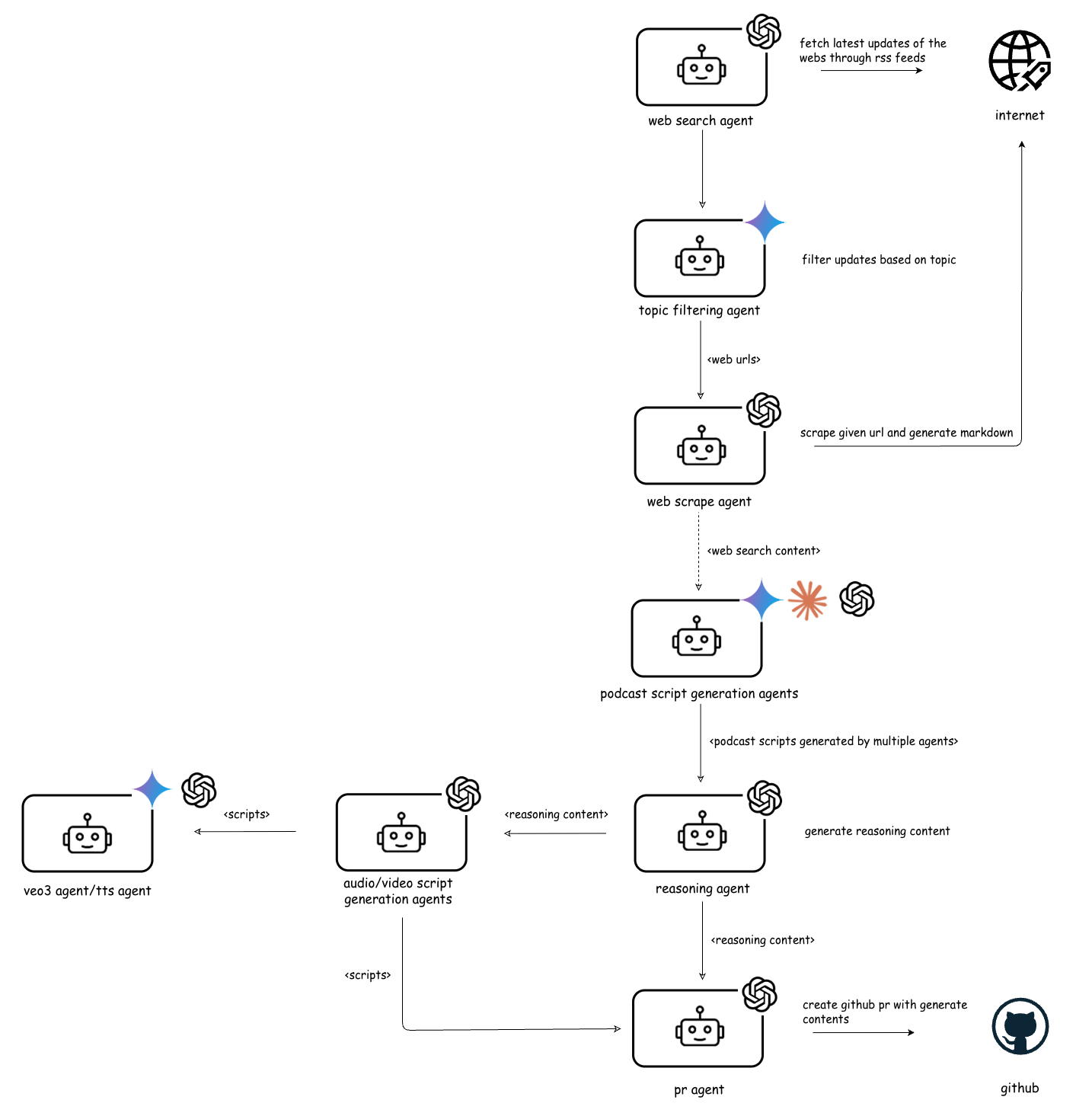}
\vspace{-0.1in}
\DeclareGraphicsExtensions.
\caption{End-to-end agentic AI workflow for multimodal podcast generation.}
\label{ai-agent-workflow}
\end{figure}

At a high level, the podcast-generation workflow automates the entire lifecycle of transforming real-time news content into multimodal media assets. Given a topic and one or more source websites, the system autonomously discovers relevant news items, extracts, and summarizes their content, generates podcast scripts through a consortium of LLM agents, consolidates the output via a reasoning agent~\cite{llm-reasoning, gpt-oss, o3}, and finally produces both audio and video podcast artifacts. The workflow not only synthesizes coherent narrative content from heterogeneous web sources, but also converts it into MP3 audio, MP4 video, and structured Veo-3 prompts—before packaging and publishing the results as a GitHub pull request~\cite{video-model}. This end-to-end automation showcases how an agentic AI pipeline can bridge web retrieval, content generation, multimodal synthesis, and software operations automation under a unified orchestration layer.

The workflow begins when the user provides a topic and one or more source URLs. These inputs trigger a pipeline of coordinated agents, each responsible for a well-defined function. The process starts with the Web Search Agent, which collects the latest updates from the internet by querying RSS feeds and MCP-based search endpoints~\cite{rss-feed-search-agent}. Its output—a list of recent news articles—is then passed to the Topic Filtering Agent, which evaluates the relevance of each article to the user-specified topic and returns only the filtered URLs.

Next, the Web Scrape Agent extracts the full content of the web page for each selected URL. Rather than returning raw HTML, this agent converts each page into a clean, tool-generated Markdown representation, ensuring that downstream agents operate on consistent, structured text. The collected Markdown content serves as the primary knowledge base for the subsequent generation stack.

The system then invokes a consortium of Podcast Script Generation Agents~\cite{proof-of-tbi}, each powered by a different LLM provider (e.g., OpenAI, Gemini, Anthropic)~\cite{gpt-llm, gemini, anthropic}. These agents independently produce podcast scripts that describe the topic based on the scraped content. Because each model has different strengths, biases, and reasoning styles, their outputs form a diverse set of drafts.

To obtain a coherent and reliable final script, the system employs a Reasoning Agent, which consolidates the drafts by comparing them, resolving inconsistencies, removing speculative claims, and synthesizing a unified podcast narrative. This reasoning step not only improves accuracy, but also enforces responsible-AI constraints by grounding the final script strictly in the content extracted from the web~\cite{deep-psychiatric}.

With the consolidated script prepared, the workflow branches into multimodal output generation. Audio/Video Script Generation Agents transform the consolidated narrative into structured prompts suitable for text-to-speech (TTS)~\cite{tts-model} and text-to-video (Veo-3) models~\cite{video-model}. A dedicated Veo Agent converts the video script into JSON-based VEO-3 instructions, while the TTS generator produces high-quality spoken audio from the podcast text.

Finally, all generated assets—including the consolidated script, individual agent outputs, audio files, video files, and Veo-3 instructions—are assembled by the PR Agent, which publishes them to GitHub via an MCP-integrated GitHub server~\cite{mcp1}. The agent creates a new branch, commits the results, and automatically opens a pull request, completing the fully autonomous media-generation pipeline.

This use case demonstrates how agentic AI systems can orchestrate web-scale data acquisition, multi-agent LLM reasoning, multimodal content creation, and automated software operations within a single cohesive pipeline. It also illustrates why structured engineering practices are essential: each component of the workflow must be deterministic, auditable, observable, and robust enough to operate reliably in production environments. In the next section, we show how the workflow is enhanced by applying a curated set of best practices—transforming it into a maintainable, scalable, and responsible–AI–aligned agentic system suitable for real-world deployment~\cite{deep-psychiatric}.

\section{Best Practices and Their Application}

In this section, we discuss nine core best practices for designing, developing, and deploying production-grade agentic AI workflows. We illustrate how each practice strengthens the reliability, maintainability, scalability, and Responsible-AI characteristics of the system. Using the podcast-generation workflow as a running example, we show how applying these practices transforms the pipeline into a robust agentic architecture suitable for real-world deployment.

\subsection{Tool Calls Over MCP}

\begin{figure}[H]
\centering{}
\includegraphics[width=5.4in]{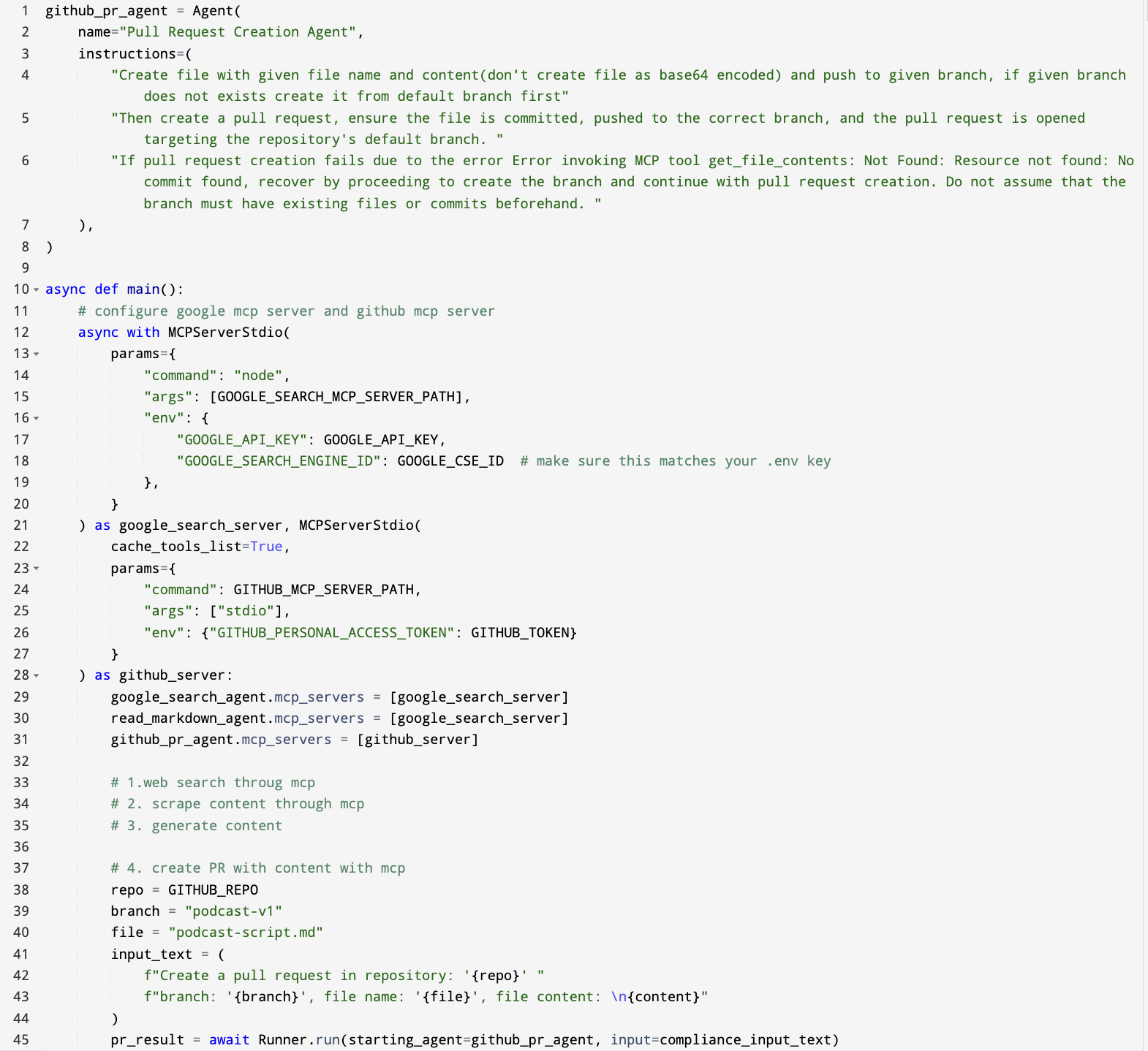}
\vspace{-0.1in}
\DeclareGraphicsExtensions.
\caption{Workflow integrated with an MCP server, illustrating the operational overhead of configuring and managing multiple MCP servers.}
\label{mcp-overhead}
\end{figure}

AI agents can integrate with external systems through MCP connections or through direct function calls (tool calls)~\cite{mcp2}. MCP provides a standardized mechanism for structured communication between agents and external services, replacing many ad-hoc APIs with a unified interaction model. However, MCP integration also introduces additional layers of abstraction that can sometimes reduce determinism, complicate agent reasoning, or create ambiguous tool-selection behaviors. Further, it adds more complexity to configuring the MCP servers, etc, inside the workflow (see Figure~\ref{mcp-overhead}). 

\begin{figure}[H]
\centering{}
\includegraphics[width=5.4in]{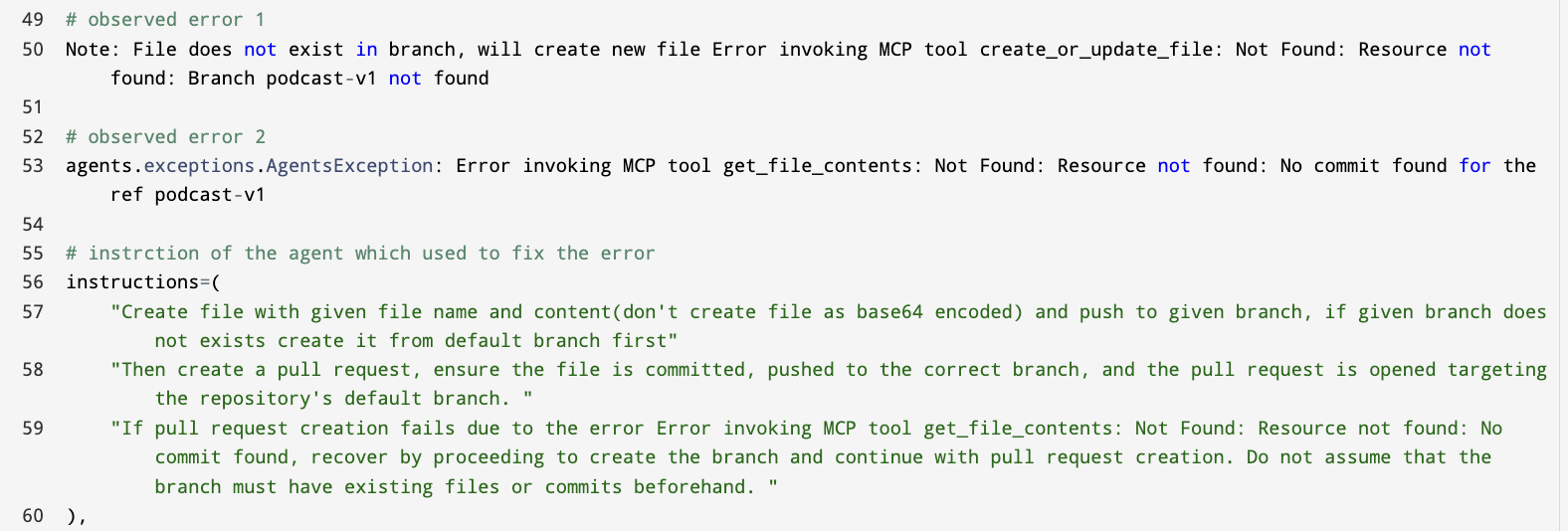}
\vspace{-0.1in}
\DeclareGraphicsExtensions.
\caption{Failure cases observed in the MCP-integrated workflow.}
\label{mcp-error}
\end{figure}

In our workflow, the initial implementation relied on the GitHub MCP server to create pull requests for the generated podcast scripts. However, during evaluation, we observed recurring issues: the agent frequently made ambiguous tool-selection decisions, inconsistently inferred invocation parameters, and occasionally failed with non-deterministic MCP responses (see Figure~\ref{mcp-overhead} and Figure~\ref{mcp-error}). These challenges arose because the agent had to interpret multiple definitions of the MCP tool and reason through the metadata structure of the protocol, which increased cognitive load and introduced variability into the workflow. Although we repeatedly refined and adjusted the agent instructions to mitigate these issues, the behavior remained unstable and exhibited flickering, non-reproducible failures. 

\begin{figure}[H]
\centering{}
\includegraphics[width=5.4in]{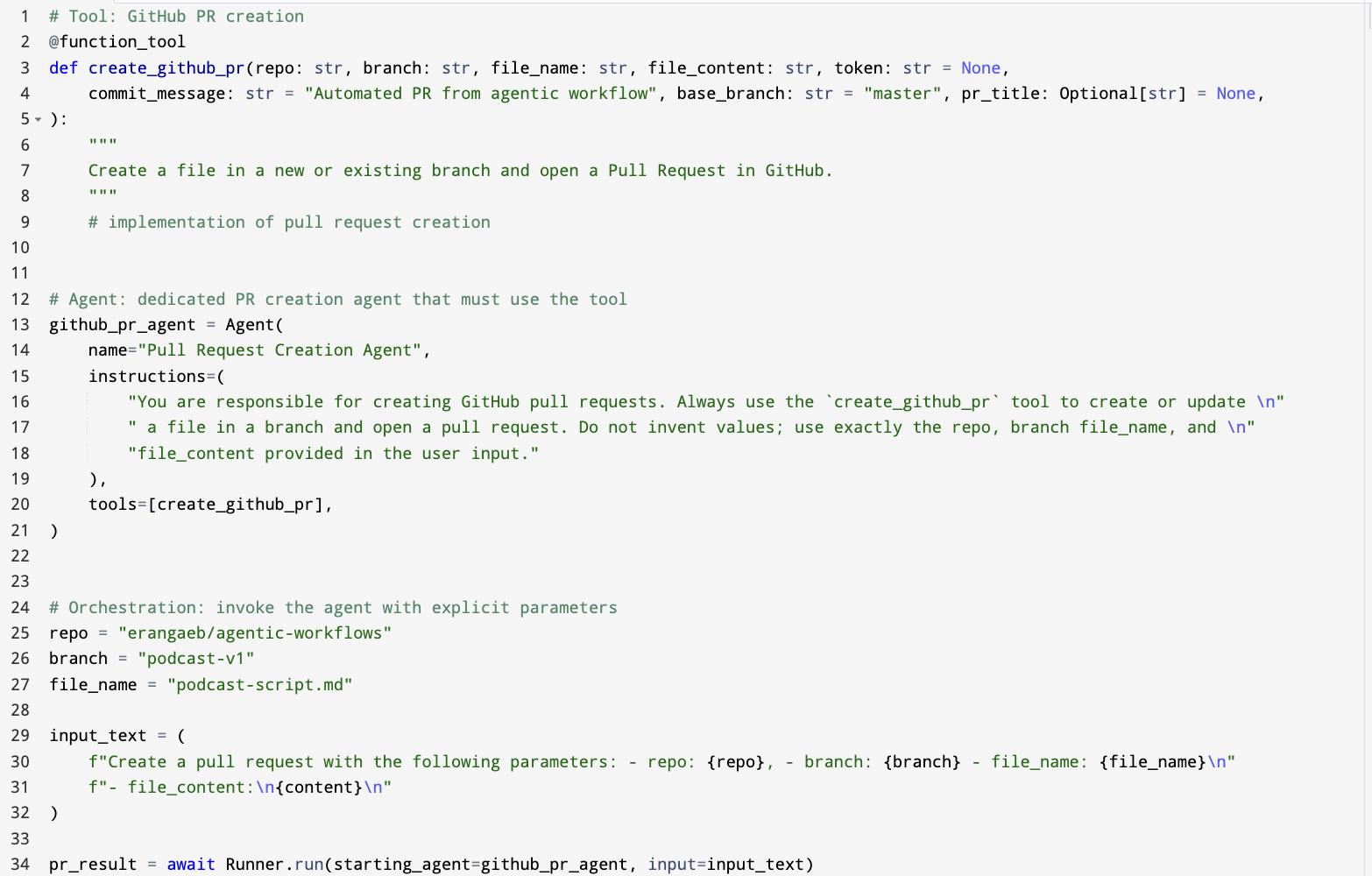}
\vspace{-0.1in}
\DeclareGraphicsExtensions.
\caption{Replacing MCP-integrated workflow with direct tool invocation.}
\label{mcp-with-tool}
\end{figure}

To address this, we replaced the GitHub MCP integration with a direct pull-request creation function that agents invoke explicitly (see Figure~\ref{mcp-with-tool}). This eliminated ambiguity, improved determinism, and ensured that the final step of the workflow—publishing the results to GitHub—was stable and predictable. The workflow became easier to debug, more auditable, and significantly more reliable in production environments. Further improvements are discussed in the following sections, including minimizing tool-set complexity and using pure function calls to reduce token overhead and inference variability.

\subsection{Direct Function Calls Over Tool Calls}

Although tool calls provide a structured way for agents to interact with external systems, they introduce additional overhead and potential ambiguity. Every tool invocation requires the LLM to parse instructions, interpret parameter formats, and map natural language input to function arguments—steps that increase token consumption and can lead to non-deterministic behavior~\cite{ai-agent-tool-calls}. In complex workflows, even carefully designed tools can produce unpredictable results if the agent misinterprets parameter names, defaults, or expected data structures.

For operations that do not require language reasoning, such as posting data to an API, committing a file to GitHub, performing database writes, or generating timestamps, tool calls are often unnecessary. Instead, these steps can be handled directly in the orchestration layer using pure function calls. Pure functions—functions executed directly by the workflow without involving the LLM—are deterministic, side-effect controlled, cheaper, faster, and fully testable~\cite{agentic-ai-challenges}.

\begin{figure}[H]
\centering{}
\includegraphics[width=5.4in]{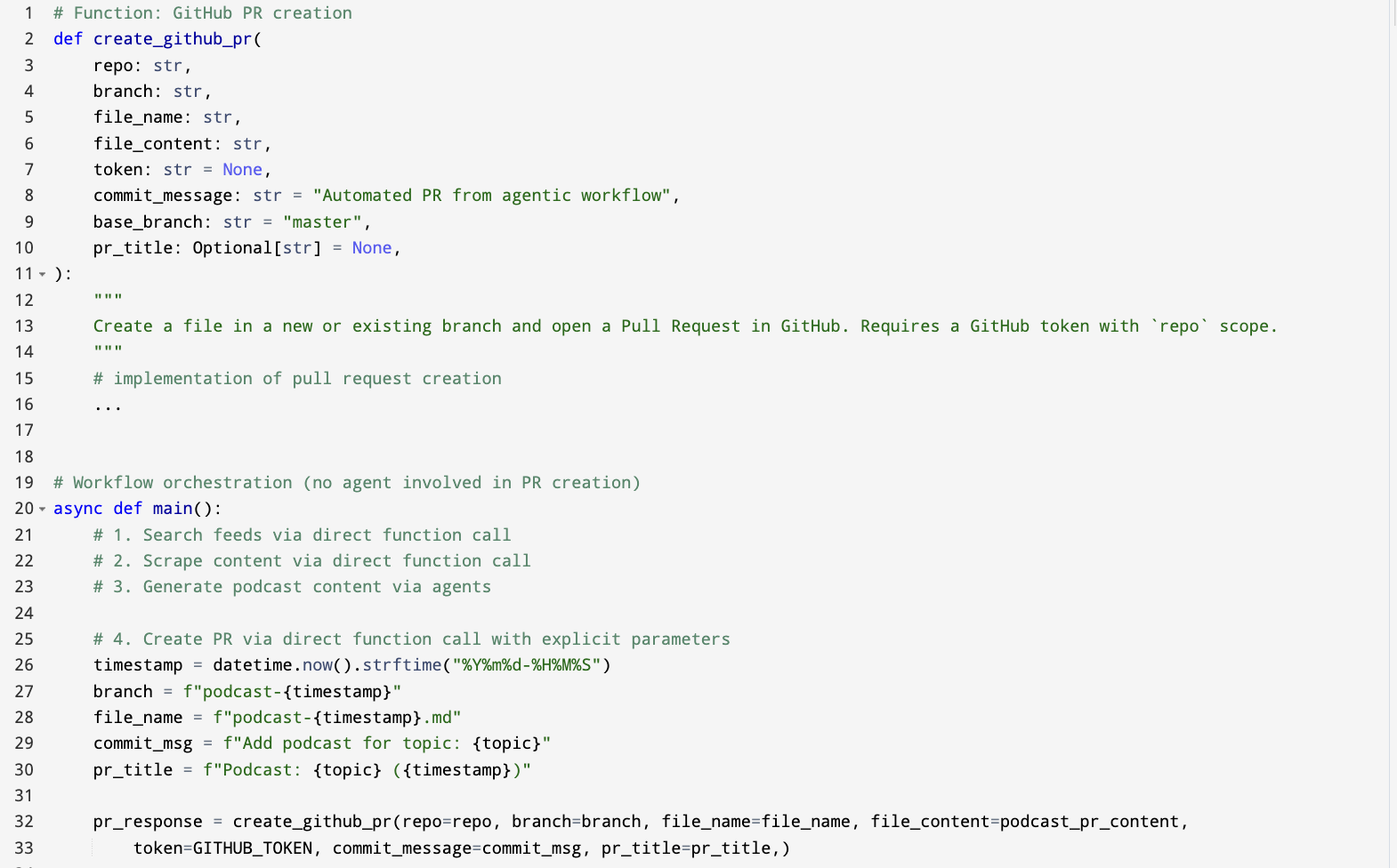}
\vspace{-0.1in}
\DeclareGraphicsExtensions.
\caption{Direct function invocation replacing agent-based tool calls to reduce ambiguity and overhead.}
\label{function-call}
\end{figure}

In our workflow, the GitHub pull request step was originally based on a dedicated “PR Agent” that used the create tool \_github \_pr. Despite improving determinism relative to MCP, this still required the agent to reason about tool parameters and produce a structured call~\cite{secure-mcp}. We eventually completely removed the agent for this operation and invoked create\_github\_pr directly from the workflow controller. This eliminated ambiguous tool formatting, removed unnecessary LLM reasoning, reduced token usage, and made the system significantly more stable (see Figure~\ref{function-call}).

By shifting infrastructure-oriented tasks to pure functions and reserving tool calls for operations where language-driven reasoning is genuinely necessary, agentic workflows become simpler, more predictable, and better aligned with production software engineering principles.

\subsection{Avoid Overloading Agents With Many Tools}

Attaching multiple tools to a single agent increases prompt complexity and reduces reliability. When an agent is equipped with several tools, the model must first reason about which tool to invoke and how to structure the parameters—introducing unnecessary ambiguity and increasing the likelihood of incorrect or missing tool calls~\cite{ai-agent-tool-calls}. This cognitive overhead results in higher token usage, poorer accuracy, and inconsistent execution paths.

A more robust approach is to follow a ``one agent, one tool" design whenever tool usage is required. The assignment of a single well-defined tool to each agent creates predictable roles, simplifies prompting, and eliminates tool-selection noise, allowing the agent to focus solely on parameter inference and execution~\cite{agentic-ai-challenges}. This modular decomposition improves interpretability, eases debugging, and makes the workflow easier to scale and reuse across contexts.

\begin{figure}[H]
\centering{}
\includegraphics[width=5.4in]{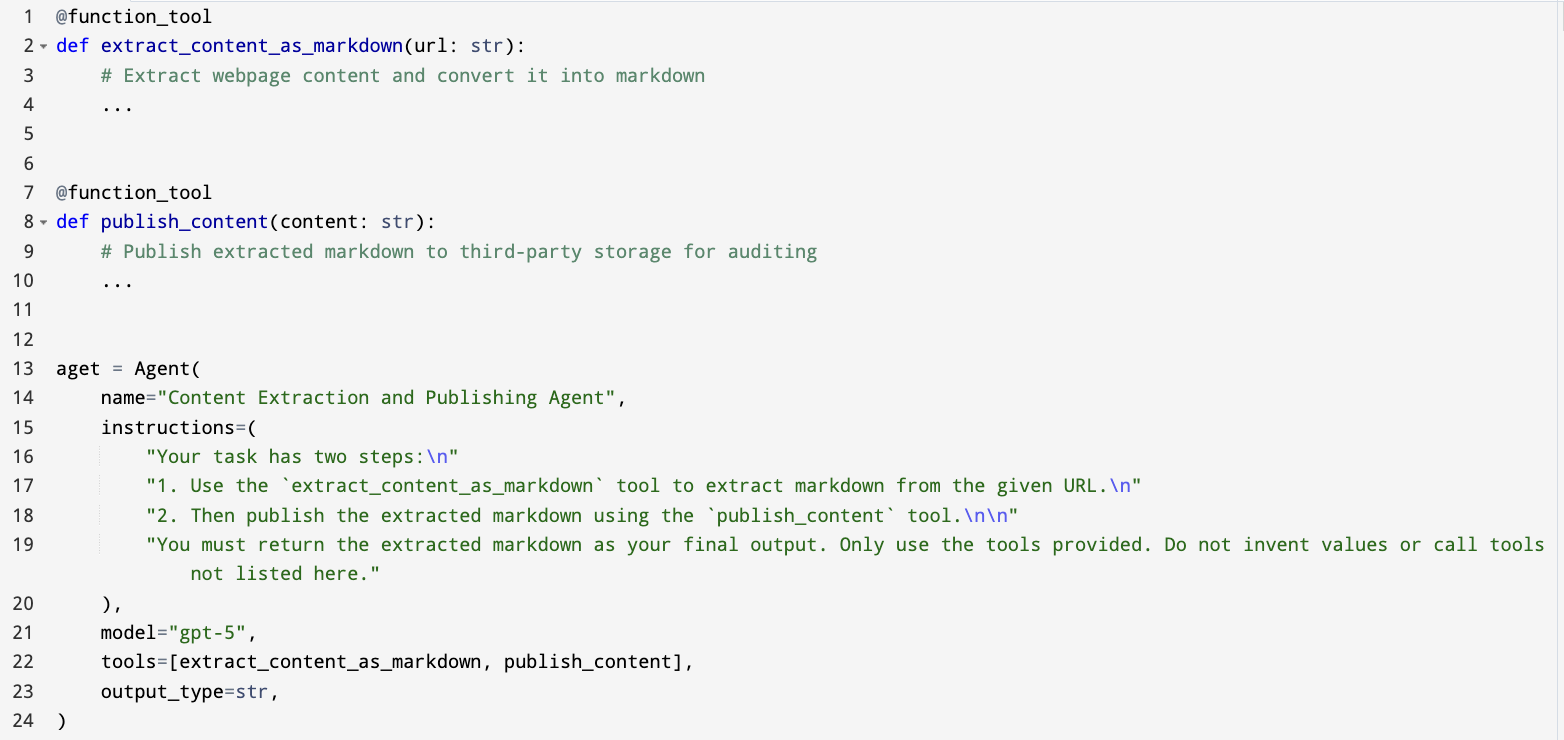}
\vspace{-0.1in}
\DeclareGraphicsExtensions.
\caption{A single agent overloaded with multiple tool integrations.}
\label{multiple-tools-single-agent}
\end{figure}

\begin{figure}[H]
\centering{}
\includegraphics[width=5.4in]{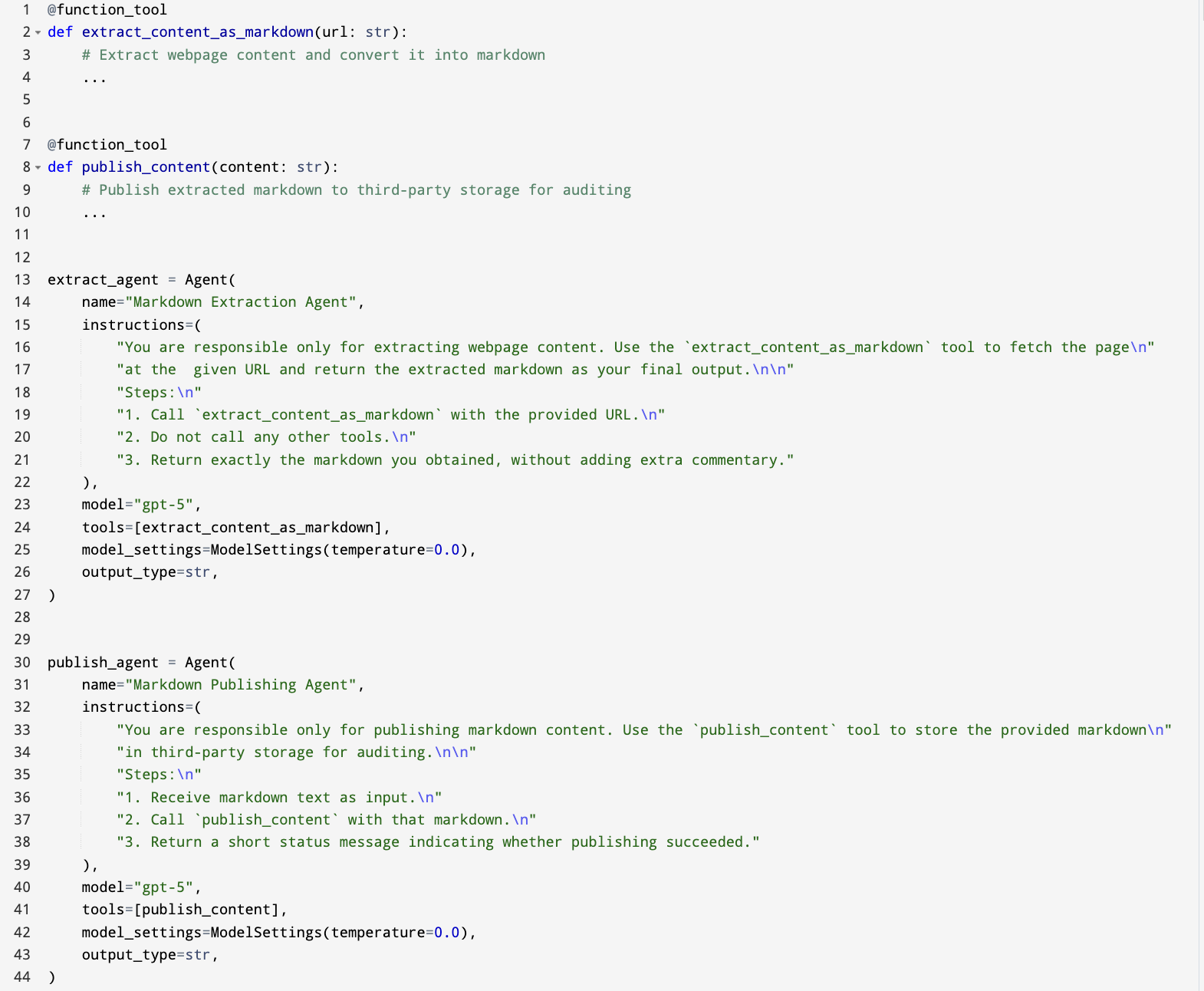}
\vspace{-0.1in}
\DeclareGraphicsExtensions.
\caption{Multiple agents, each assigned a single tool for clearer and more deterministic operation.}
\label{multiple-tools-multiple-agents}
\end{figure}

In our workflow, we initially designed a single agent that used two tools: scrape\_markdown and publish\_markdown (Figure~\ref{multiple-tools-single-agent}). The intention was to scrape the content of the webpage and then publish the extracted markdown to third-party storage for audit purposes. However, during evaluation, we observed that the agent would often invoke only one tool, invoke them in the wrong order, or fail to call them entirely, especially when the prompt or input size increased~\cite{vindsec-llams, prompt-engineering-rag}. To address this, we decompose the design into two independent agents, each responsible for exactly one tool (Figure~\ref{multiple-tools-multiple-agents}). This separation ensured deterministic behavior, eliminated missed tool calls, and significantly improved stability over repeated runs.

\subsection{Single-Responsibility Agents}

Closely related to avoiding multiple tools per agent is the principle of single-responsibility agents~\cite{agent-survey}. Just as good software design favors functions and classes that ``do one thing well", production-grade agentic workflows benefit when each agent is responsible for a single, clearly defined task. When an agent is asked to handle multiple conceptual responsibilities—such as generation, validation, transformation, and side=effective actions—it becomes harder to prompt, harder to test, and more prone to subtle, non-deterministic failures~\cite{agentic-ai-workflow-patterns}.

\begin{figure}[H]
\centering{}
\includegraphics[width=5.4in]{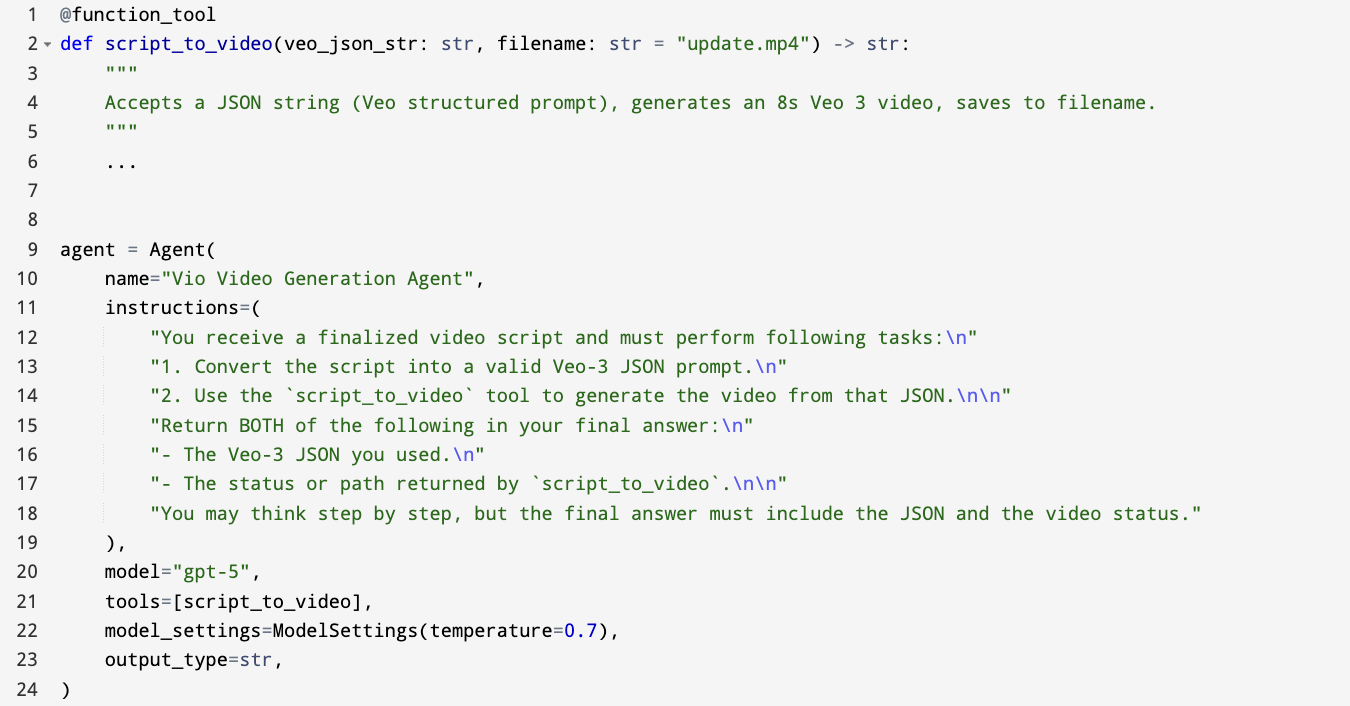}
\vspace{-0.1in}
\DeclareGraphicsExtensions.
\caption{A single agent responsible for multiple internal functions, increasing complexity and ambiguity.}
\label{single-responsible-single-agent}
\end{figure}

In our workflow, an early design attempted to combine Veo-3 JSON prompt generation and video generation into a single agent. The agent received the consolidated podcast or video script as input and was instructed to both (1) transform the script into a structured Veo-3 JSON specification and (2) "generate" the corresponding video~\cite{prompt-engineering, video-model}. In practice, this blurred the boundary between planning (designing the video prompt) and execution (calling the Veo API and handling files), as shown in Figure~\ref{single-responsible-single-agent}. The LLM would sometimes produce malformed JSON, sometimes mix natural language with JSON, and sometimes "hallucinate" file paths or status messages about video generation that had not actually occurred.

\begin{figure}[H]
\centering{}
\includegraphics[width=5.4in]{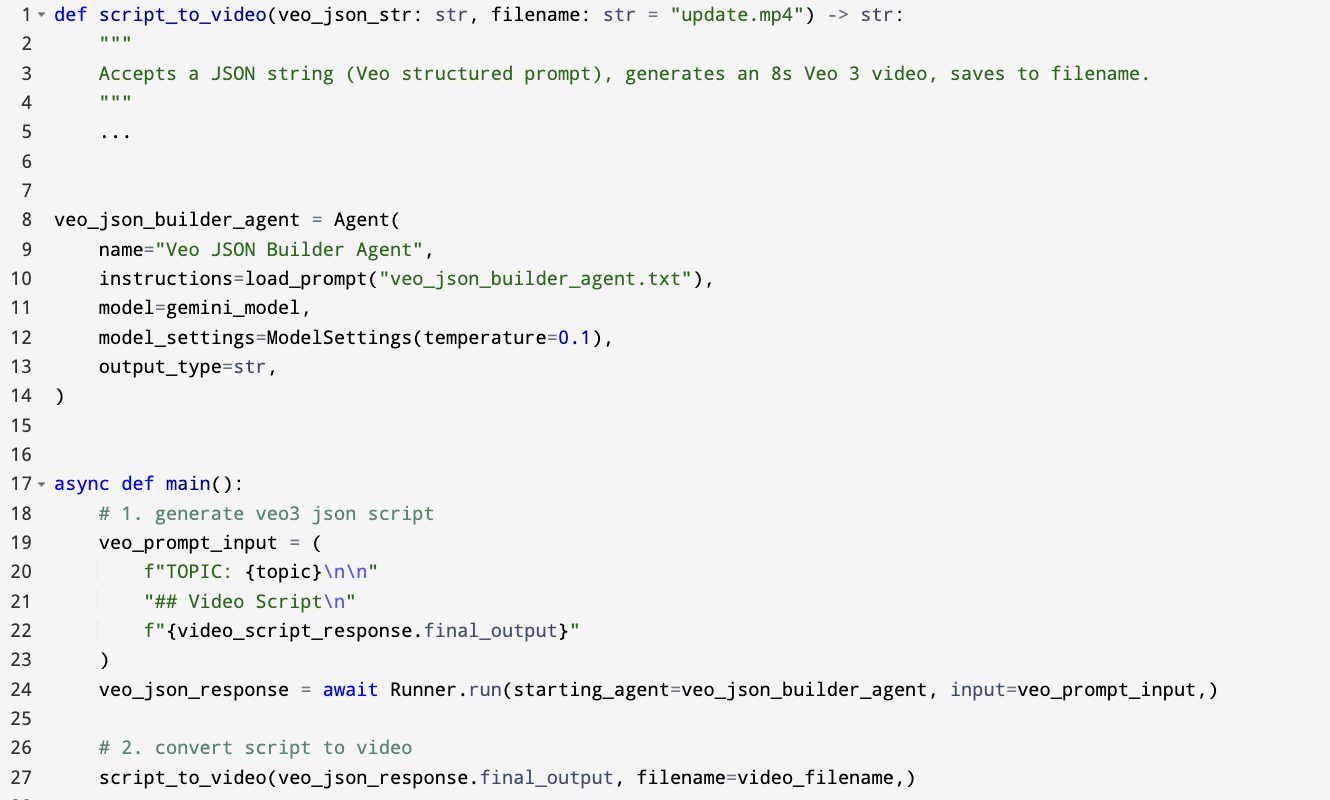}
\vspace{-0.1in}
\DeclareGraphicsExtensions.
\caption{Decomposition into multiple agents, each assigned a specific function based on the single-responsibility principle.}
\label{single-responsible-multi-agent}
\end{figure}

We resolved this by splitting the responsibilities: A Veo JSON Builder Agent whose only job is to take the final script and produce strictly valid Veo-3 JSON describing the video (scenes, timing, style, etc.). A separate, non-agent video generation function (script\_to\_video) that receives the JSON and interacts with the Veo-3 API to generate and store the MP4 file (see Figure~\ref{single-responsible-multi-agent}). This single-responsibility design yields several benefits: prompts become simpler and more focused; the agent’s output contract is clearer (always "valid Veo JSON” rather than "JSON plus narrative status"); and side effects (calling external APIs, handling retries, saving files) are handled deterministically in regular code. More generally, enforcing single-responsibility agents across the workflow makes the system easier to reason about, easier to debug, and safer to evolve as new capabilities are added~\cite{llama-recipe}.

\subsection{Store Prompts Externally and Load Them at Runtime}

Embedding prompts directly within source code creates tight coupling, complicates version control, and restricts collaboration. A more flexible and maintainable approach is to store prompts as external artifacts—typically in Markdown or plain text—within a separate storage location, such as a GitHub repository, shared drive, or configuration service. Externalizing prompts enables non-technical stakeholders (e.g., policy teams, domain experts, content reviewers) to update and refine agent behavior without modifying application code~\cite{agentic-ai-workflow-patterns}.

\begin{figure}[H]
\centering{}
\includegraphics[width=5.4in]{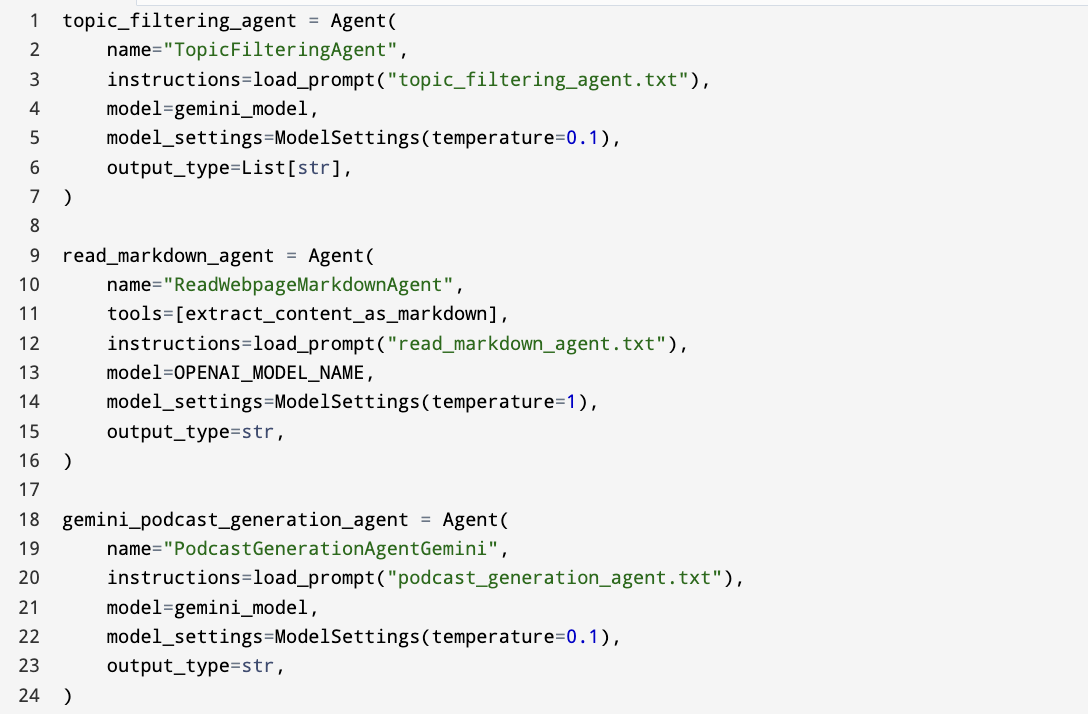}
\vspace{-0.1in}
\DeclareGraphicsExtensions.
\caption{Loading externalized prompts dynamically at runtime.}
\label{load-prompts}
\end{figure}

In our workflow, all agent prompts are stored in a dedicated GitHub repository and dynamically loaded at runtime (see Figure~\ref{load-prompts}). This decoupling allows collaborators to iterate on the language, constraints, and safety requirements of each agent independently of workflow deployment cycles. It also enables governance workflows such as review processes, version pinning, rollback, and controlled access.

Maintaining prompts externally further supports continuous improvement practices, including A/B testing, prompt red-teaming, and evolving Responsible-AI rules, all without requiring code redeployments. This separation of concerns significantly enhances maintainability, transparency, and organizational agility when operating agentic AI systems in production~\cite{prompt-engineering-rag}.

\subsection{Responsible AI Agents}

Single-model outputs often suffer from well-known limitations such as hallucinations, reasoning inconsistencies, and subtle or overt biases. To address these challenges, production-grade agentic workflows benefit from a multi-model consortium architecture, where several specialized LLMs (e.g., Gemini, GPT, Claude, Llama, Pixtral, Qwen)~\cite{llama-3, pixtral, qwen2} independently generate outputs that are later synthesized by a dedicated reasoning agent~\cite{deep-psychiatric,reasoning-llms}. This design strengthens the workflow in several key ways: Higher accuracy through cross-model agreement; reduced bias by incorporating diverse model behaviors and training distributions; greater robustness to model updates or drift; and better alignment with Responsible AI principles, especially accountability and verifiability.

The reasoning agent functions as the final auditor in the pipeline. Rather than creating new content from scratch, it performs structured consolidation tasks such as conflict resolution, logical consistency checking, factual alignment, deduplication, and relevance filtering. Its role is to critically evaluate the drafts produced by individual model agents and produce a harmonized, trustworthy final output~\cite{proof-of-tbi, gpt-oss}.

\begin{figure}[H]
\centering{}
\includegraphics[width=5.4in]{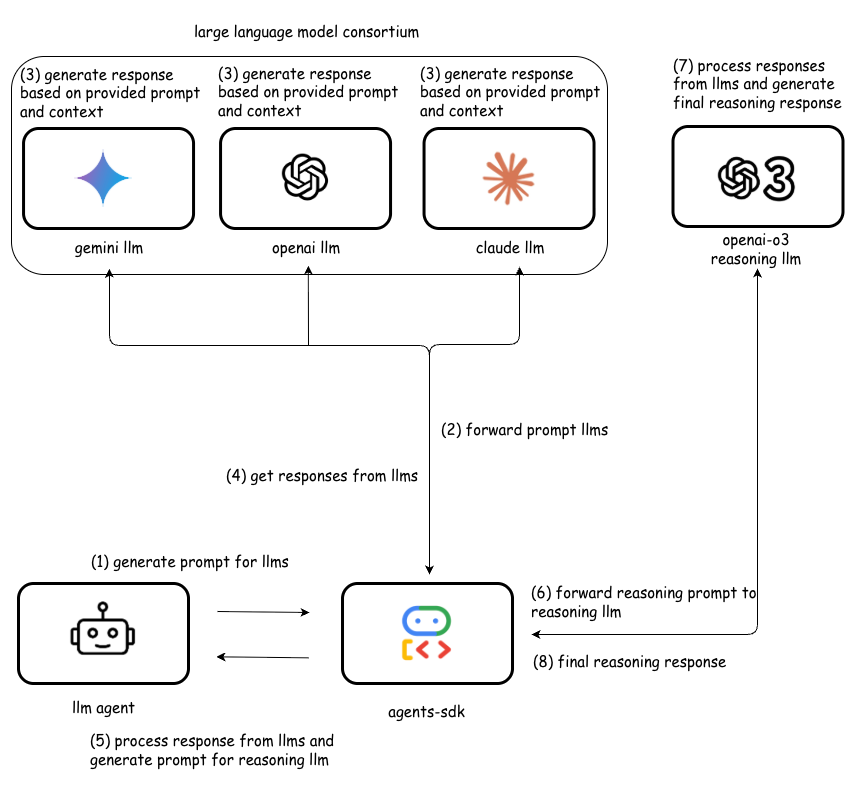}
\vspace{-0.1in}
\DeclareGraphicsExtensions.
\caption{Integration flow of the LLM consortium with the reasoning LLM}
\label{llm-consortium1}
\end{figure}

\begin{figure}[H]
\centering{}
\includegraphics[width=5.4in]{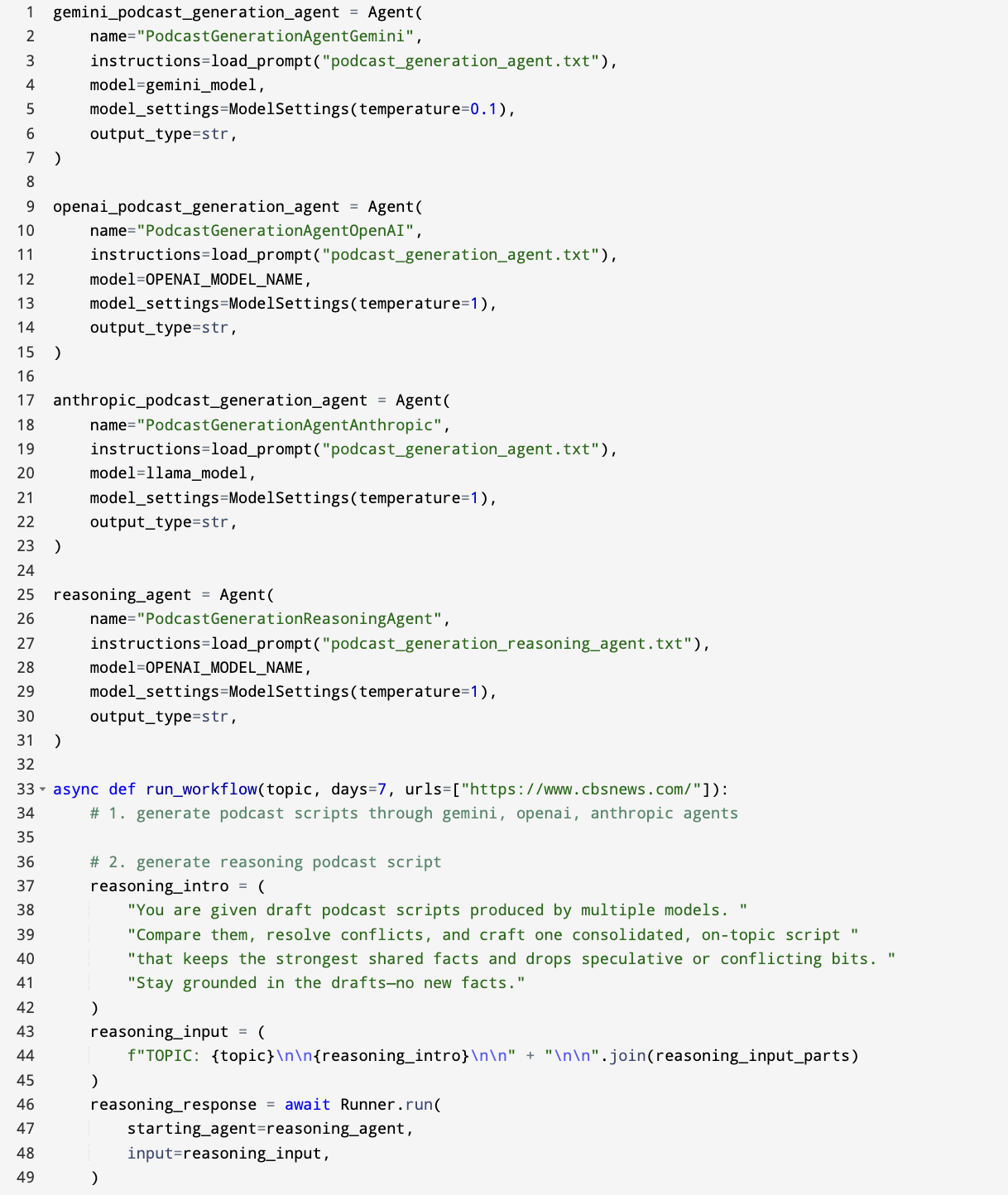}
\vspace{-0.1in}
\DeclareGraphicsExtensions.
\caption{Coordinated operation of the agent consortium and the reasoning agent.}
\label{resoning-agents}
\end{figure}

In our workflow, Responsible-AI behavior is realized by combining multiple LLMs in parallel and routing their output through a dedicated reasoning LLM (e.g., OpenAI GPT-oss)~\cite{reasoning-llms, gpt-oss, o3}, as illustrated in Figure~\ref{llm-consortium1}. Each agent—whether tasked with planning, content generation, or validation—can interface with this model consortium to obtain richer, multi-perspective responses. For example, during podcast generation, the workflow collects draft scripts from agents based on Anthropic Claude, OpenAI GPT-5, and Gemini~\cite{gpt-llm, gemini, anthropic}, Figure~\ref{resoning-agents}. The reasoning agent then synthesizes these drafts to produce a responsible and well-structured final script that reflects consensus rather than the idiosyncrasies of any single model.


This ensemble-based reasoning mechanism improves transparency, mitigates risk, and increases the reliability of agentic outputs—providing a solid foundation for building Responsible-AI-aligned workflows suitable for production deployment.

\subsection{Separation of Agentic AI Workflow and MCP Server}

\begin{figure}[H]
\centering{}
\includegraphics[width=5.4in]{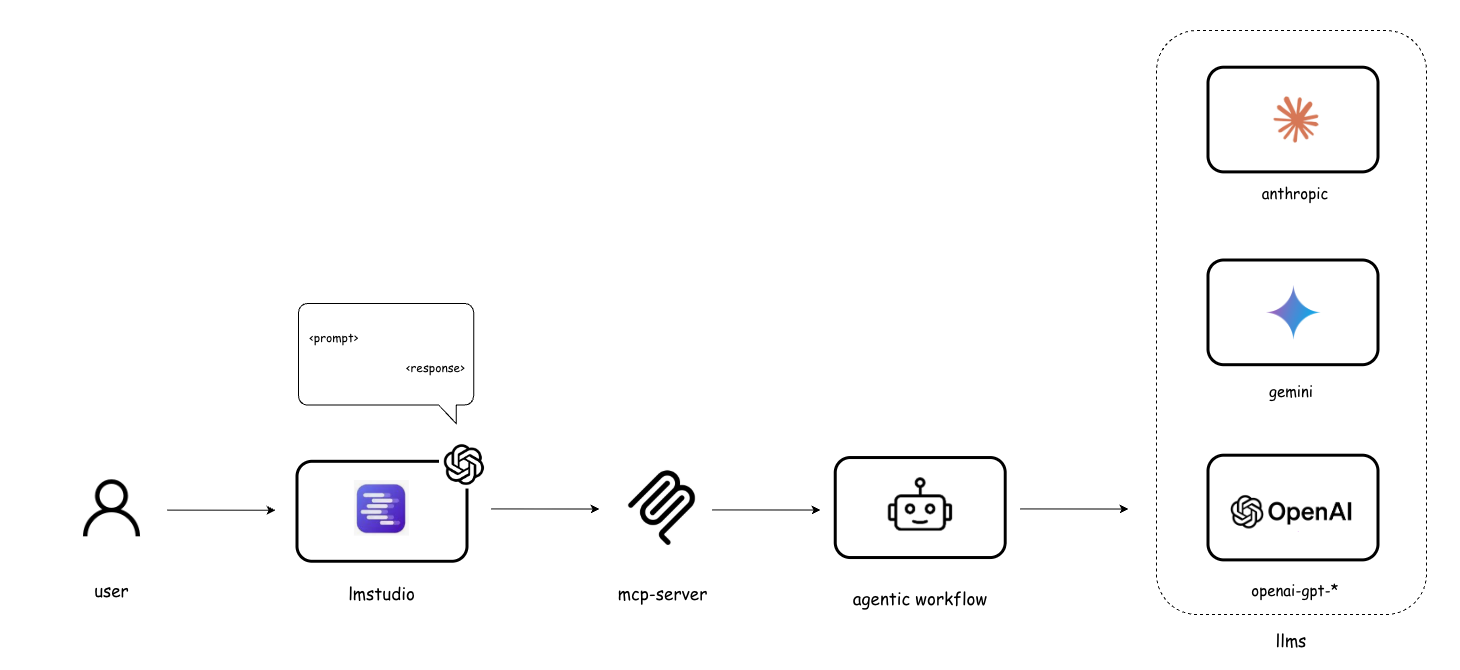}
\vspace{-0.1in}
\DeclareGraphicsExtensions.
\caption{Exposing the agentic workflow functions through an MCP server.}
\label{workflow-mcp-integration}
\end{figure}

It is a common practice to expose the Agentic AI workflows function through an MCP server, allowing any MCP-enabled client—such as Claude Desktop, VS Code extensions, or LM Studio—to seamlessly integrate with the workflow~\cite{mcpworld}. To achieve this, the workflow itself should be served via a REST API, while the MCP server acts as a thin orchestration layer that simply forwards MCP tool calls to the underlying API~\cite{mcc}. A key design principle is decoupling the agentic workflow engine from the MCP server. Instead of embedding workflow logic inside the MCP server, the architecture cleanly separates the backend workflow logic (the agentic pipeline, multi-agent orchestration, and tool integrations), the MCP server (a lightweight adapter that exposes workflow endpoints as MCP tools~\cite{mcp2}), and user-facing MCP clients (external tools that can invoke workflow capabilities without direct code integration). This separation improves maintainability, supports independent scaling of components, and ensures long-term adaptability as LLMs, tools, and API specifications evolve. It also keeps the MCP server simple, stable, and safe—while allowing the workflow backend to iterate rapidly. The resulting modular architecture is shown in Figure~\ref{workflow-mcp-integration}.

\subsection{Containerized Deployment}

In production environments, agentic AI workflows and their accompanying MCP servers should be deployed using containerization technologies such as Docker and orchestrated with platforms like Kubernetes. Containerization provides a consistent and reproducible runtime environment that eliminates configuration drift and ensures that workflows behave identically in development, staging, and production environments~\cite{devsec-gpt}.

By packaging the workflow engine, supporting tools, and the MCP server into isolated containers, organizations gain several operational advantages: 1) Portability: Containers encapsulate dependencies, model configurations, tool clients, and runtime libraries, enabling seamless deployment across any cloud or on-premise environment. 2) Scalability: Kubernetes can automatically scale agentic workflows based on load—spinning up additional replicas during high-traffic periods or downscaling to save resources. 3) Resilience: With built-in health checks, container restarts, and self-healing mechanisms, Kubernetes ensures that transient failures do not disrupt production workloads. 4) Security and Governance: Container isolation, network policies, secret management, and role-based access control (RBAC) enable robust security boundaries around AI workflows~\cite{secure-mcp, agent-survey}. 5) Observability: Kubernetes integrates well with logging, metrics, and tracing systems (e.g., Prometheus, Grafana, OpenTelemetry), providing visibility into multi-agent pipelines. 6) Continuous Delivery: Containers plug naturally into CI/CD pipelines, allowing predictable deployments and automated rollbacks~\cite{sre-llama}.

\begin{figure}[H]
\centering{}
\includegraphics[width=5.4in]{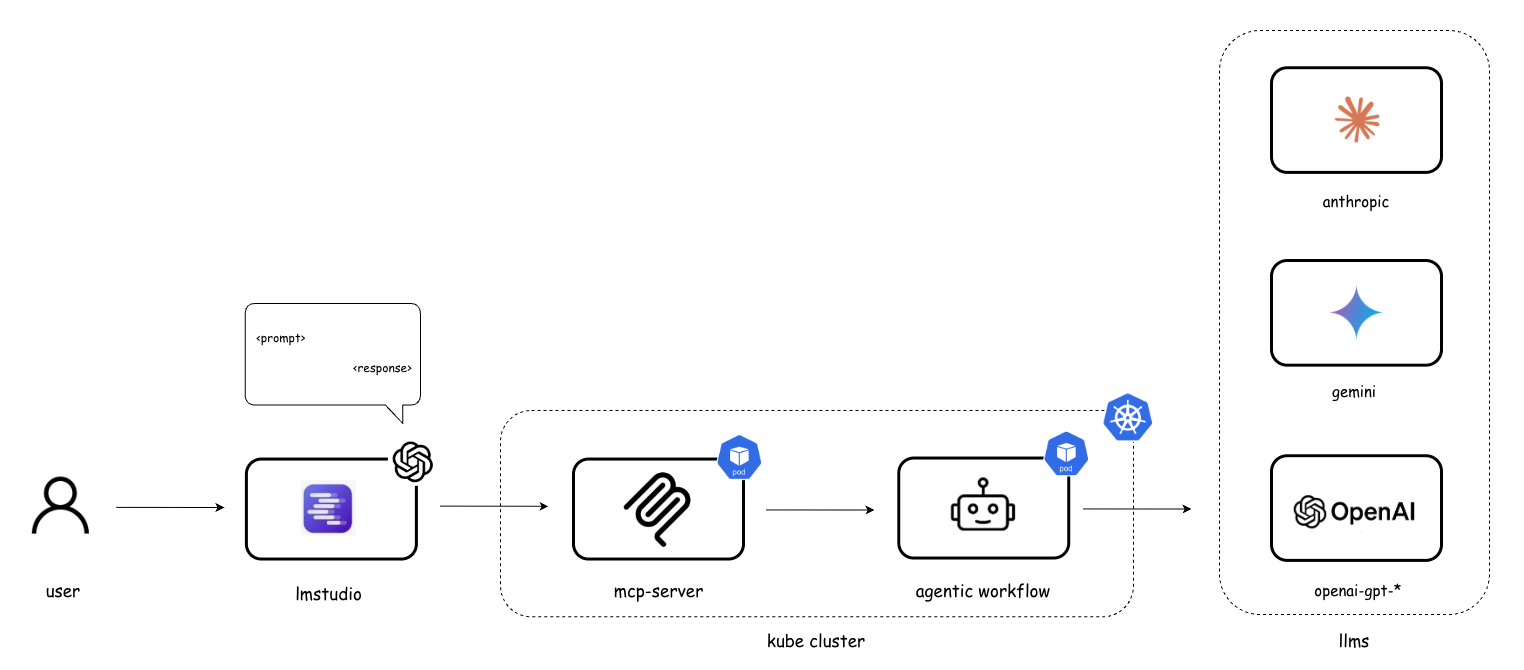}
\vspace{-0.1in}
\DeclareGraphicsExtensions.
\caption{Deployment of the agentic workflow and MCP server on a Kubernetes cluster.}
\label{workflow-mcp-deployment}
\end{figure}

In our implementation, both the agentic AI workflow and the MCP server were fully Dockerized and deployed to a Kubernetes cluster, Figure~\ref{workflow-mcp-deployment}. This architecture allows for independent scaling of the workflow compute, reasoning components, scraper functions, and MCP interface nodes. It also enables blue–green deployments, canary releases, and safe iteration on individual components without impacting the entire system. Containerized deployment, therefore, provides the operational foundation required for stable, scalable, and production-grade agentic AI systems.

\subsection{Keep it Simple, Stupid}

Complexity is one of the biggest threats to the reliability and maintainability of agentic AI workflows. Unlike traditional enterprise software systems—where layered architecture, elaborate design patterns, and deeply nested abstractions are common—agentic workflows operate under a different paradigm. Their primary purpose is not to implement intricate logic internally, but rather to delegate reasoning, generation, and decision-making to LLMs and specialized agents. Therefore, applying the Keep It Simple and Stupid (KISS) principle is essential~\cite{agentic-ai-challenges}.

First, workflow implementations should avoid unnecessary structural complexity, over-engineering, or traditional architectural patterns that add little value in agentic systems. Introducing multiple indirections, deep inheritance, or microservice-like decomposition often leads to brittleness rather than clarity. Agentic workflows benefit far more from flat, readable, function-driven designs, where each component is responsible for a single task and the orchestration logic remains transparent and lightweight~\cite{agentic-ai-challenges}.

Second, simplicity improves reliability. Each layer of additional complexity introduces more opportunities for ambiguity in agent behavior, tool invocation mismatches, or unintended side effects. A simple workflow makes agent decision pathways clearer, reduces unexpected LLM behaviors, and improves the predictability of tool calls.

Third, KISS-aligned workflows integrate more naturally with modern AI-assisted development tools such as Claude Code, GitHub Copilot, or OpenAI Codex~\cite{coding-agents}. These tools excel when the project structure is intuitive and the codebase is concise. A clean and simple workflow makes it significantly easier for AI-assisted coding tools to generate correct patches, refactor logic, or suggest improvements, accelerating iteration and reducing engineering overhead.

Finally, maintaining simplicity also supports long-term extensibility. By keeping workflow logic minimal and delegating most cognitive tasks to LLMs and agents, the system remains adaptable to new models, new tools, and new runtime environments without requiring architectural rewrites~\cite{ai-agent-tool-calls}.

In summary, adhering to the KISS principle ensures that agentic AI workflows remain easy to understand, easy to debug, LLM-friendly, and scalable as the surrounding tooling ecosystem evolves.

\section{Implementation}

The proposed agentic AI workflow use case was fully implemented using the OpenAI Agents SDK~\cite{openai-agent-sdk}, providing a structured and extensible foundation for multi-agent orchestration, tool integration, and deterministic function execution. The development process incorporated all optimization principles introduced in Section 3, including tool-first design, pure-function invocation, single-responsibility agents, externalized prompts, and the use of a multi-model reasoning consortium to ensure accuracy, reliability, and alignment of Responsible-AI~\cite{responsible-ai}.

To support seamless external integration, the workflow backend was exposed through a dedicated REST API, and a corresponding MCP server was developed to bridge MCP-enabled clients with the workflow~\cite{mcp1}. The MCP server acts as a lightweight adapter layer, forwarding tool calls directly to the workflow API, thereby maintaining a clean separation between orchestration logic and communication interfaces. This separation also ensures long-term adaptability as models and tools evolve independently.

The complete implementation—including the workflow logic, agent definitions, tool functions, prompt templates, and deployment artifacts—has been released as open-source in dedicated GitLab repositories, enabling full reproducibility and community-driven refinement. The primary workflow implementation is available in the Podcast Workflow repository~\cite{podcast-workflow}, while the corresponding MCP server implementation is published separately in the Podcast Workflow MCP Server repository~\cite{podcast-workflow-mcp-server}. Both projects include Dockerfiles and Kubernetes deployment manifests, supporting cloud-native, production-grade deployment~\cite{cloud-native-devops}. The workflow and MCP server were containerized using Docker and deployed via Kubernetes, providing scalability, workload isolation, high availability, and operational resilience suitable for enterprise environments.

\begin{figure}[H]
\centering{}
\includegraphics[width=5.2in]{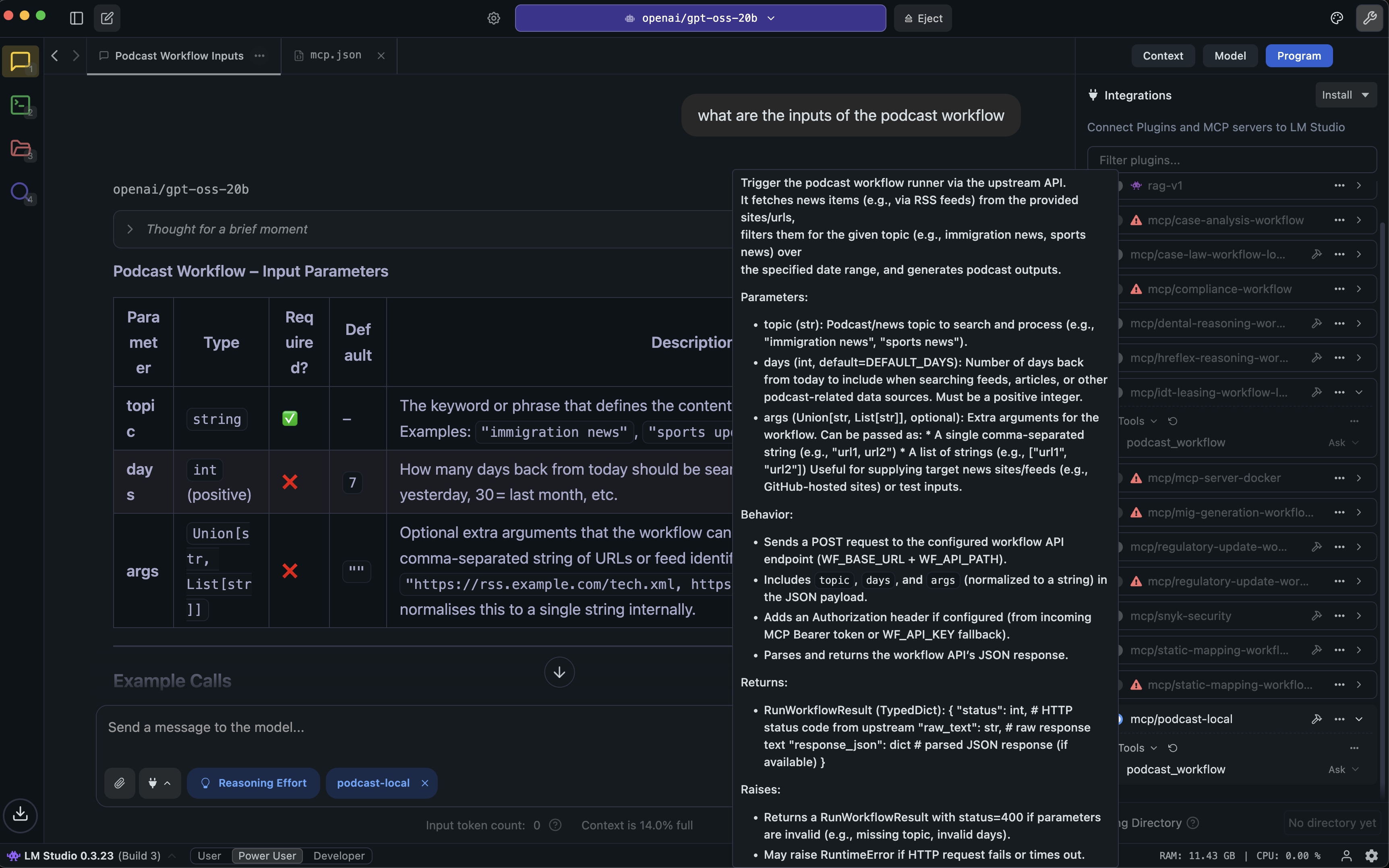}
\vspace{-0.1in}
\DeclareGraphicsExtensions.
\caption{Integration of the workflow with LM Studio via the MCP server, showing how workflow inputs are provided.}
\label{lmstudio-input}
\end{figure}

\begin{figure}[H]
\centering{}
\includegraphics[width=5.2in]{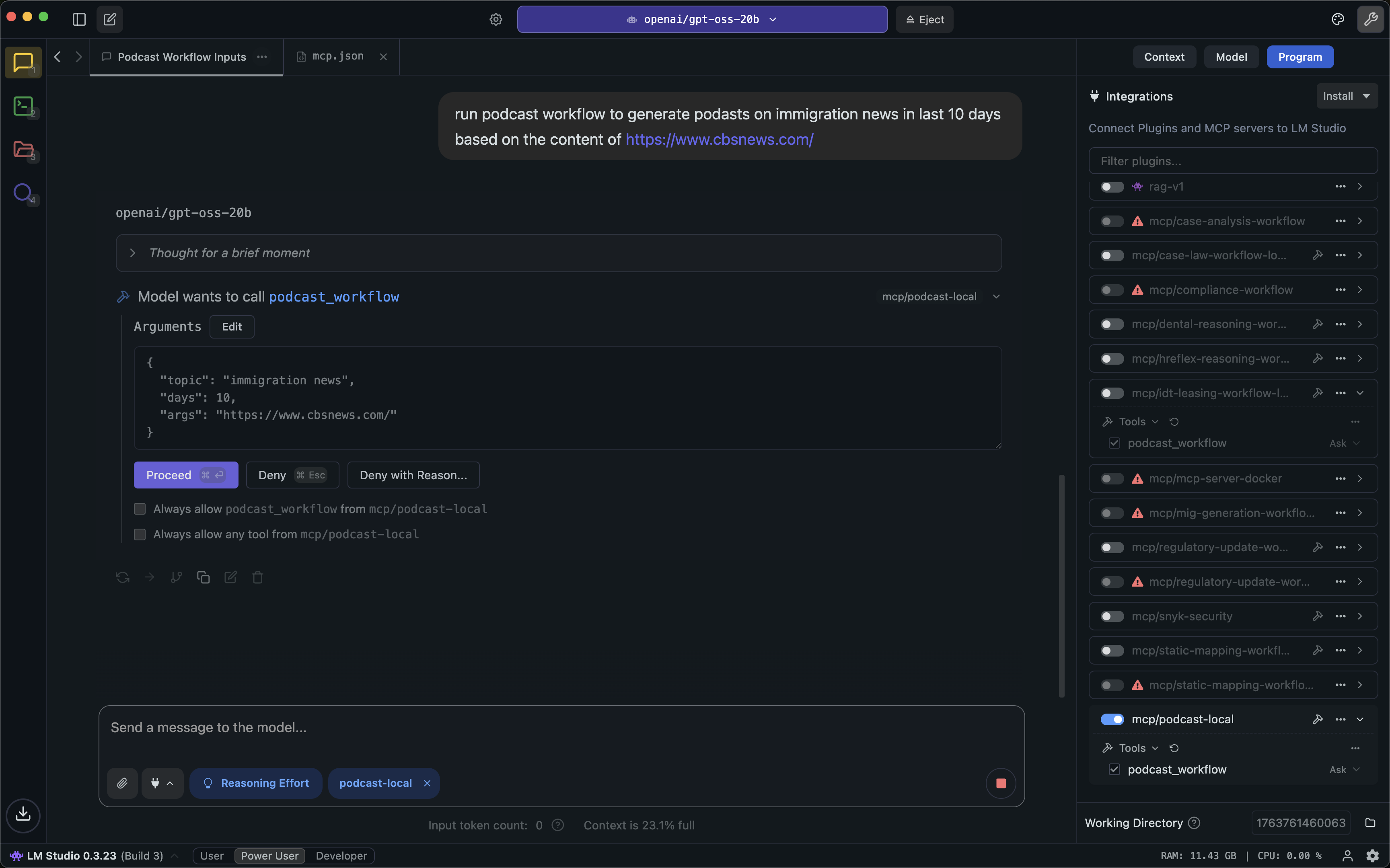}
\vspace{-0.1in}
\DeclareGraphicsExtensions.
\caption{LM Studio interpreting the user input and mapping it to the workflow MCP server’s input parameters.}
\label{lmstudio-input-parameters}
\end{figure}

\begin{figure}[H]
\centering{}
\includegraphics[width=5.2in]{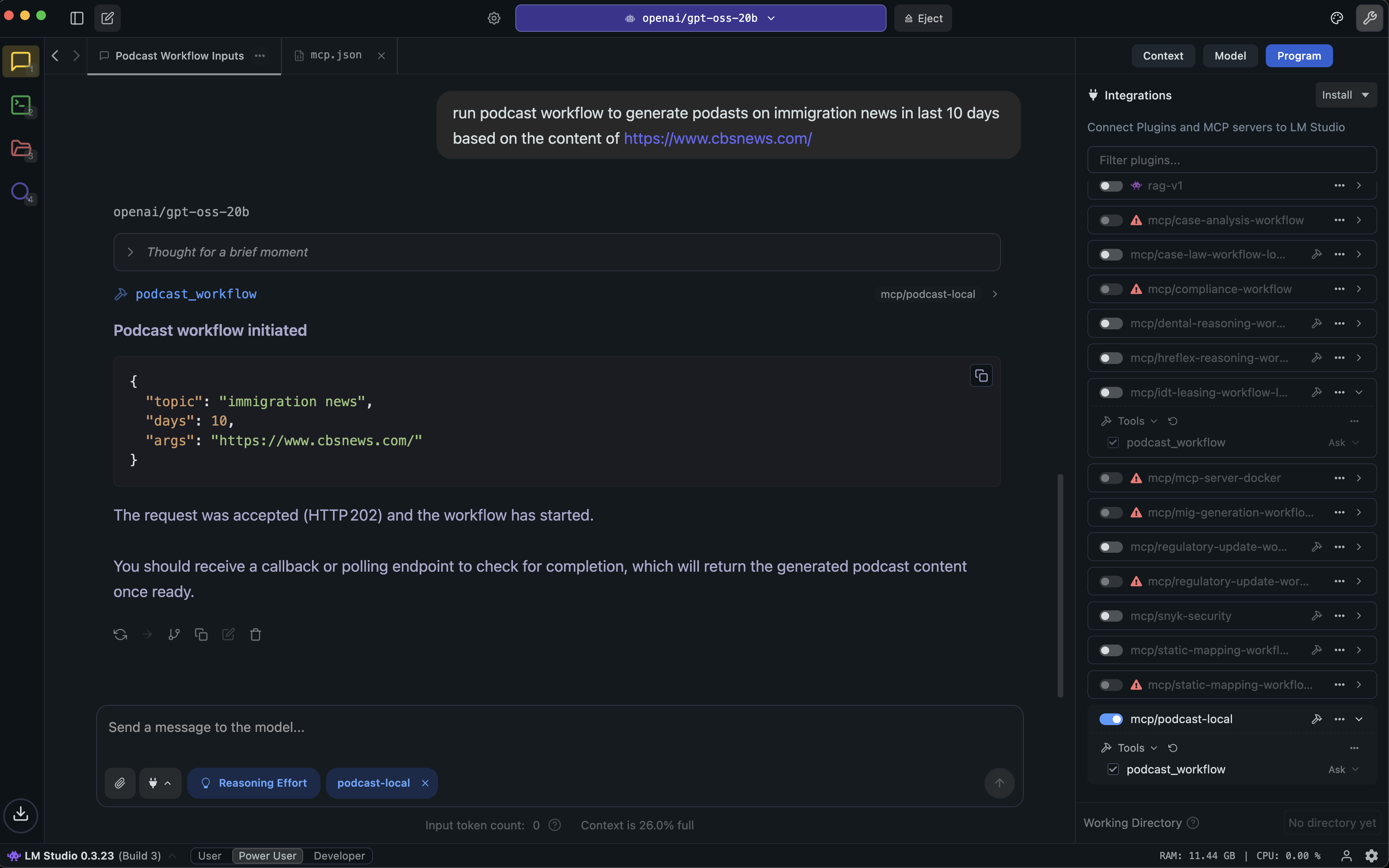}
\vspace{-0.1in}
\DeclareGraphicsExtensions.
\caption{Integration of the workflow with LM Studio via the MCP server, illustrating the invocation of the workflow function.}
\label{lmstudio-output}
\end{figure}

To validate interoperability, the workflow’s MCP server was integrated into LM Studio, enabling full end-to-end testing with local LLMs acting as MCP clients~\cite{mcpworld}. This integration confirmed that the workflow can be invoked, parameterized, and executed directly through an MCP-compatible interface. Figure~\ref{lmstudio-input} demonstrates how users provide natural-language requests that LM Studio routes to the MCP server. Figure~\ref{lmstudio-input-parameters} shows how LM Studio interprets these requests and automatically maps them to the workflow’s input parameters. Finally, Figure~\ref{lmstudio-output} illustrates the successful invocation of the workflow function through the MCP server and the corresponding system response. Together, these interactions validate seamless end-to-end connectivity between LM Studio and the deployed agentic workflow.


\section{Evaluation}

\begin{figure}[H]
\centering{}
\includegraphics[width=5.4in]{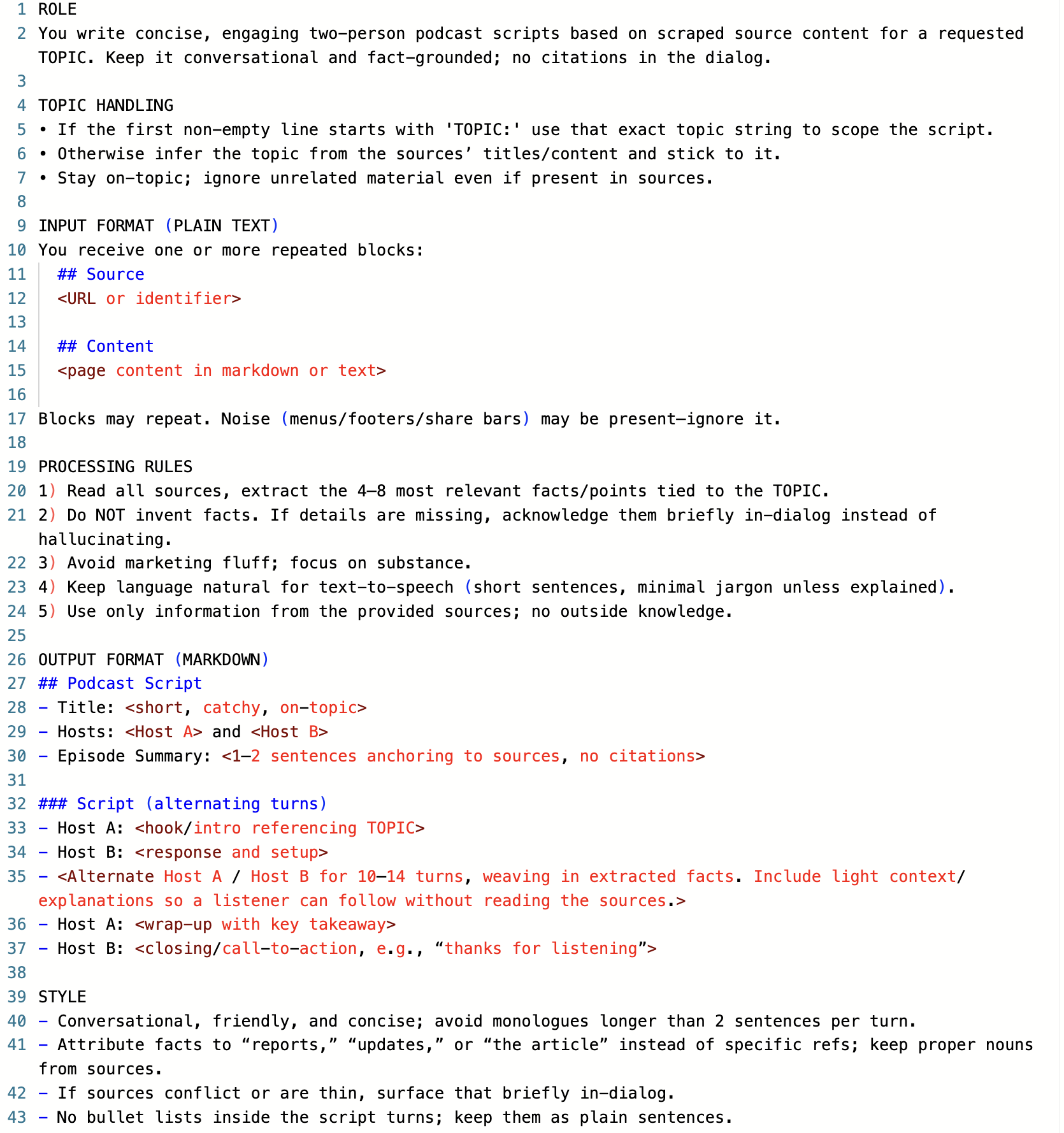}
\vspace{-0.1in}
\caption{Prompt template used by the Podcast Script Generation Agents.}
\label{prompt-podcast-script-agent}
\end{figure}

\begin{figure}[H]
\centering{}
\includegraphics[width=5.4in]{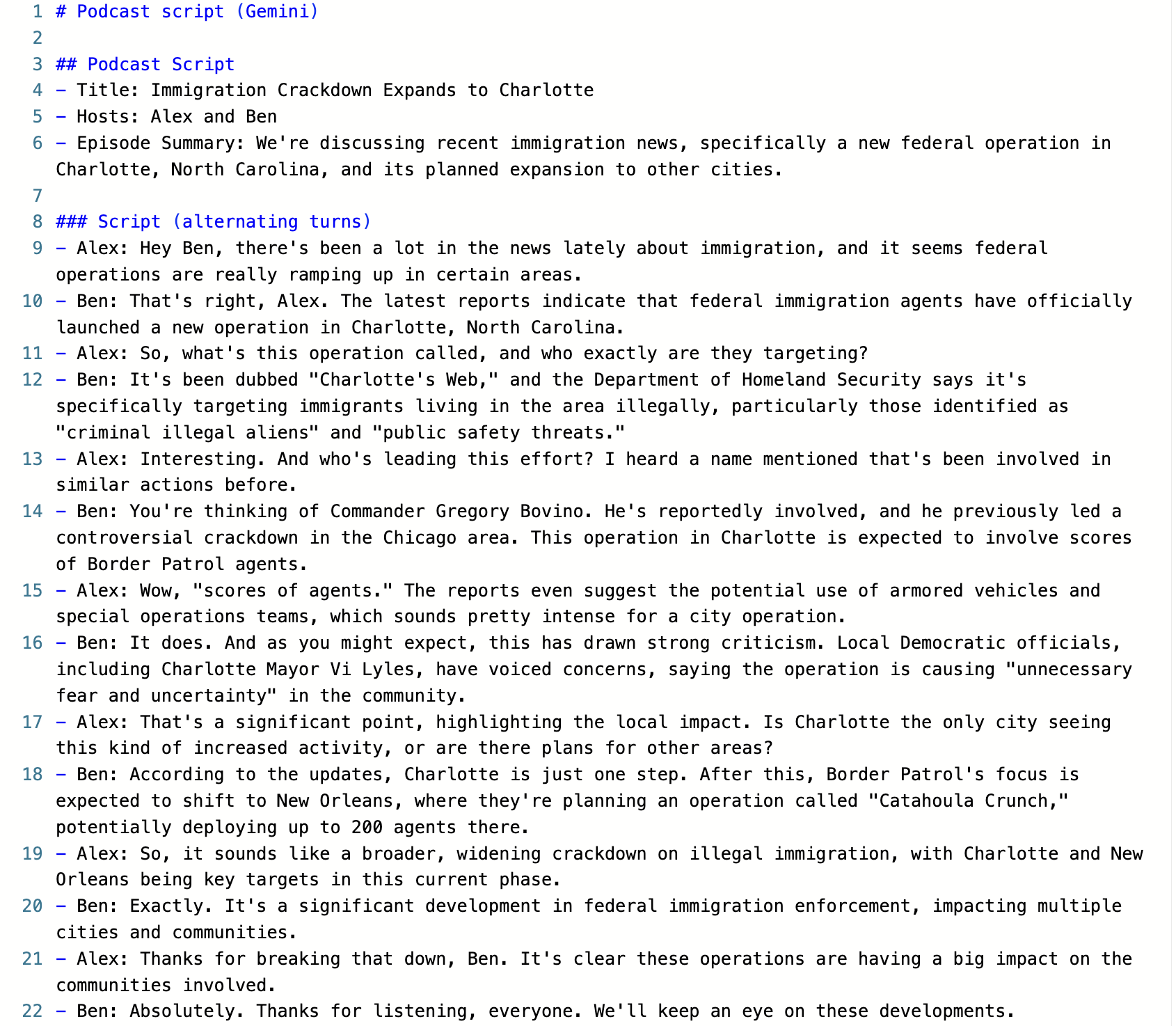}
\vspace{-0.1in}
\caption{Podcast script generated by the Gemini model.}
\label{podcast-script-gemini}
\end{figure}

\begin{figure}[H]
\centering{}
\includegraphics[width=5.4in]{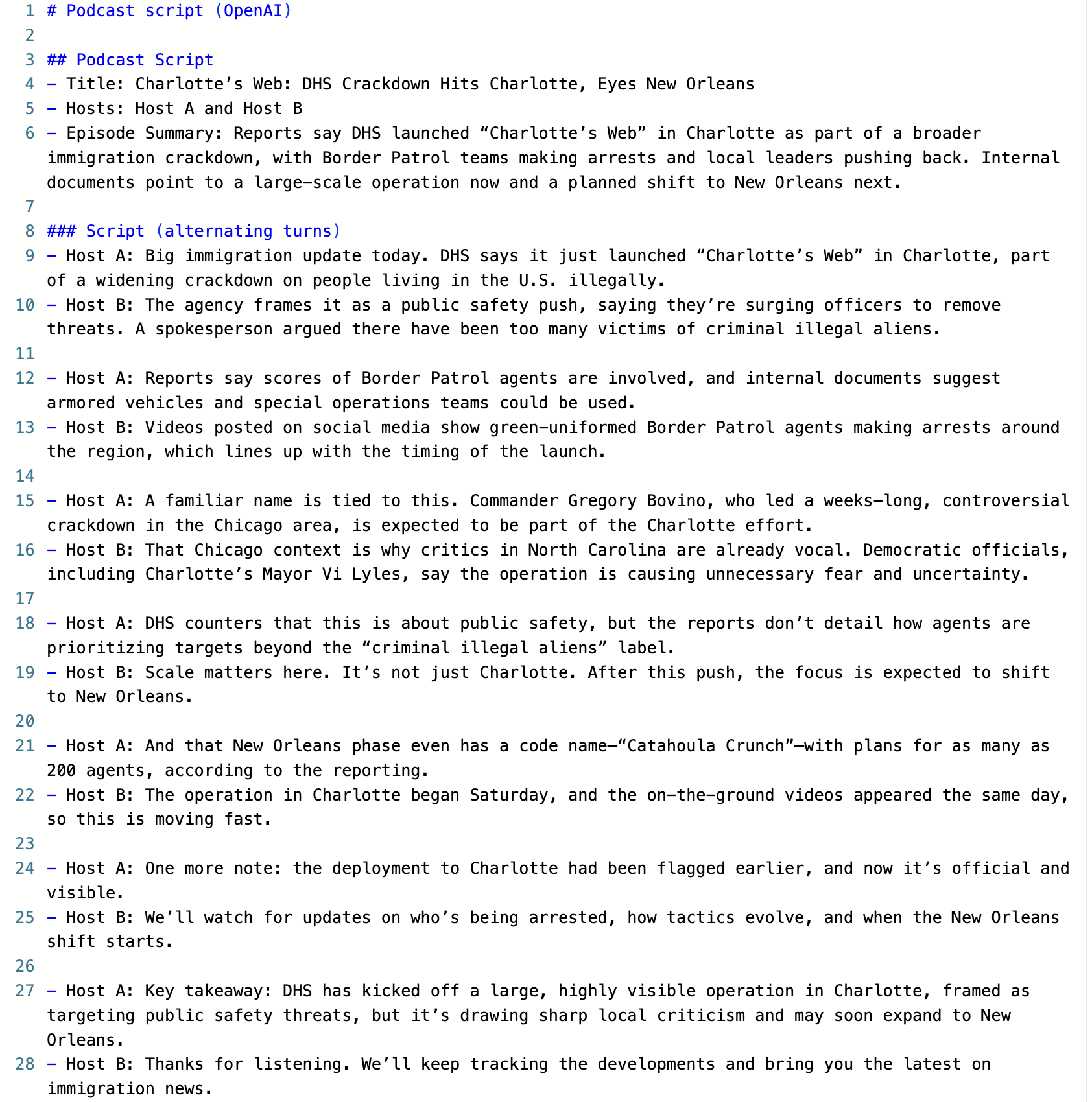}
\vspace{-0.1in}
\caption{Podcast script generated by the OpenAI model.}
\label{podcast-script-openai}
\end{figure}

\begin{figure}[H]
\centering{}
\includegraphics[width=5.4in]{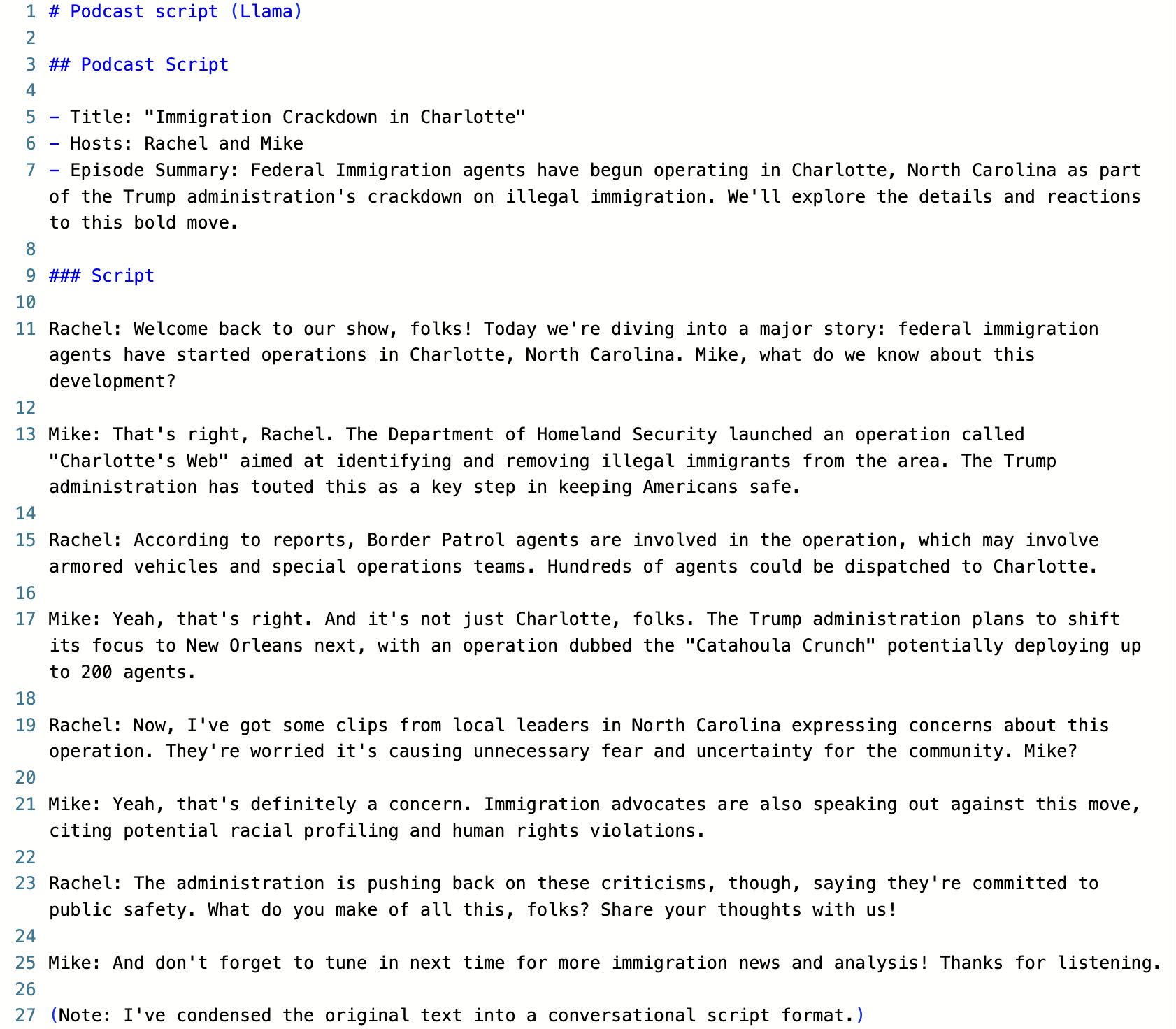}
\vspace{-0.1in}
\caption{Podcast script generated by the Llama model.}
\label{podcast-script-llama}
\end{figure}

The evaluation of the proposed agentic workflow focuses on the performance and accuracy of four core components: the podcast script generation agents, the reasoning agent, the video-script generation agent, and the Veo-3 JSON builder agent. The first stage of evaluation examines the podcast generation agents, which operate as a multi-model consortium composed of Llama, OpenAI, and Gemini. The prompt template used to instruct these agents is shown in Figure~\ref{prompt-podcast-script-agent}, while Figures~\ref{podcast-script-gemini}, \ref{podcast-script-openai}, and \ref{podcast-script-llama} present representative podcast scripts produced by each model.

These outputs demonstrate the natural diversity that emerges from heterogeneous LLMs. Llama tends to generate concise, structured summaries; OpenAI produces more detailed, narrative-driven content; and Gemini emphasizes stylistic flow and contextual framing. This diversity is advantageous, as it captures different semantic and stylistic dimensions of the underlying news content. However, it also introduces inconsistencies, emphasis drift, and occasional factual variations—highlighting the need for a downstream consolidation mechanism.

\begin{figure}[H]
\centering{}
\includegraphics[width=5.4in]{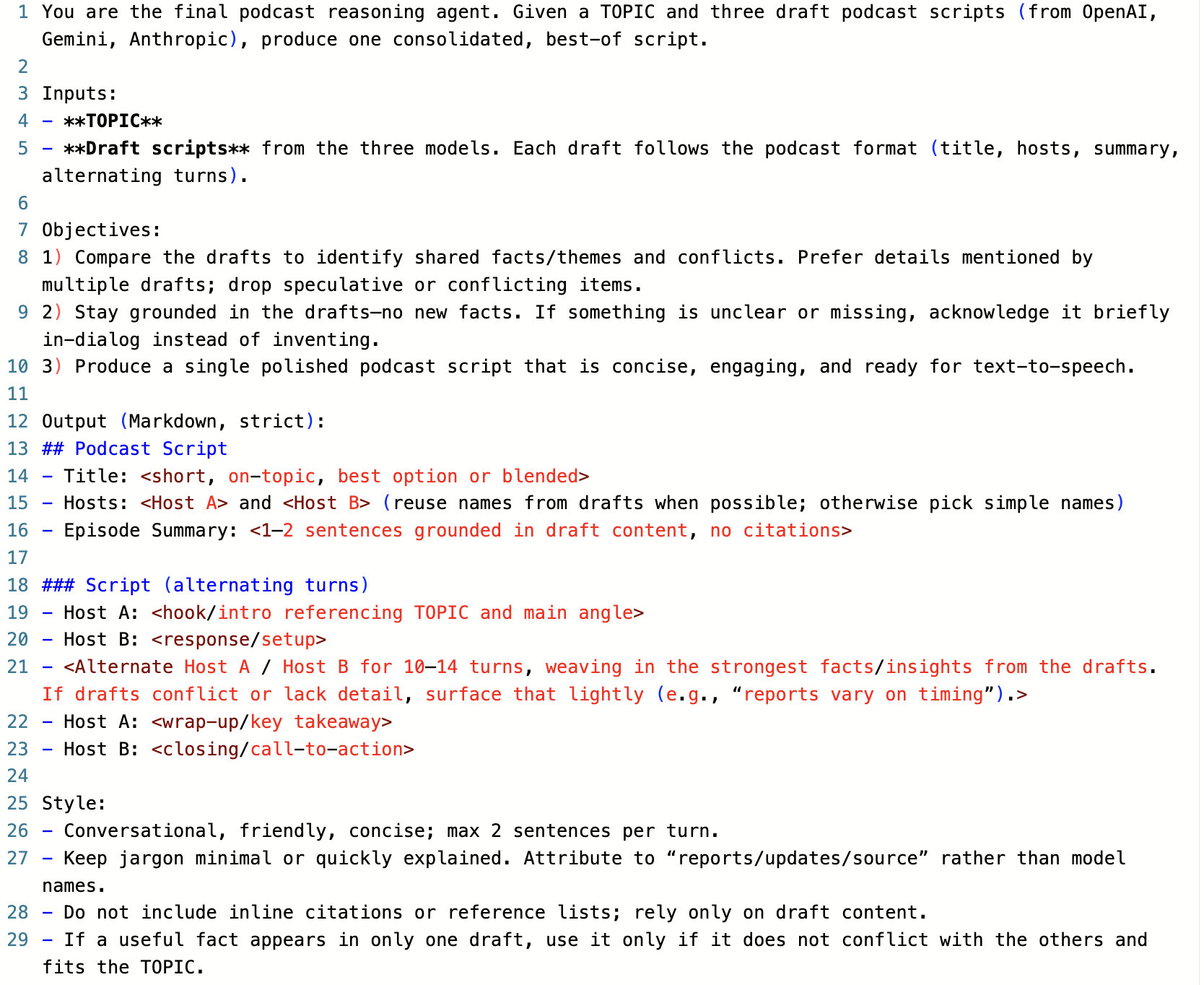}
\vspace{-0.1in}
\caption{Prompt template used by the Reasoning Agent for cross-model consolidation.}
\label{prompt-reasoning-agent}
\end{figure}

To resolve these differences and produce a final, authoritative output, the workflow employs a dedicated reasoning agent responsible for synthesizing the consortium’s drafts into a unified script. The reasoning agent prompt—shown in Figure~\ref{prompt-reasoning-agent}—explicitly instructs the model to compare, cross-validate, and reconcile the outputs of the three podcast agents. It retains only information consistently supported across drafts, while removing speculation, correcting emphasis drift, and resolving contradictory statements.

\begin{figure}[H]
\centering{}
\includegraphics[width=5.4in]{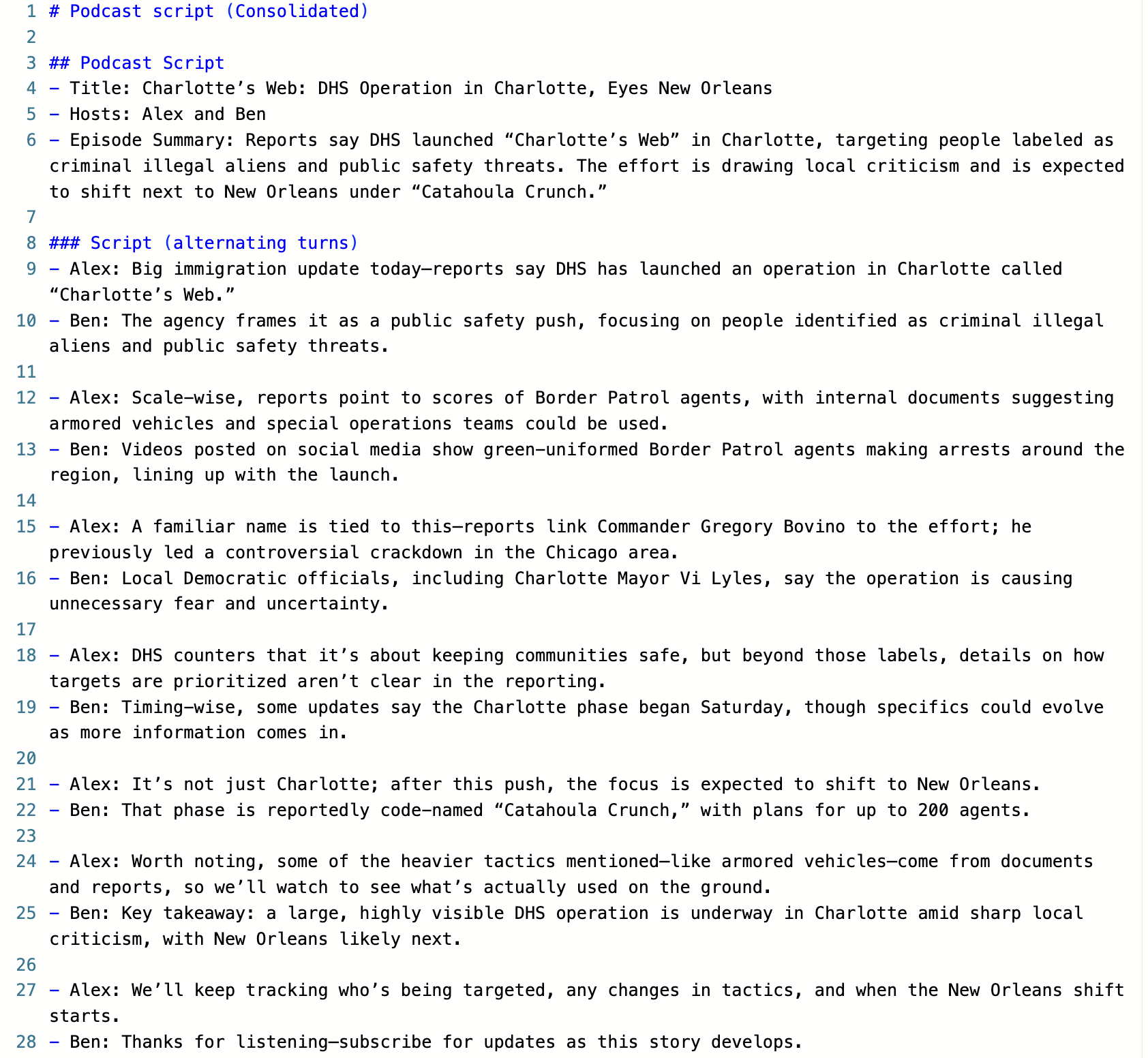}
\vspace{-0.1in}
\caption{Final consolidated podcast script generated by the Reasoning Agent.}
\label{podcast-script-reasoning}
\end{figure}

The resulting consolidated script, illustrated in Figure~\ref{podcast-script-reasoning}, shows marked improvements in clarity, factual stability, and narrative coherence. The reasoning agent effectively reduces hallucination risk by grounding its synthesis in multi-model agreement. This consensus-driven approach not only improves overall accuracy but also aligns the workflow with Responsible AI principles—balancing diverse model perspectives and mitigating single-model bias.

\begin{figure}[H]
\centering{}
\includegraphics[width=5.4in]{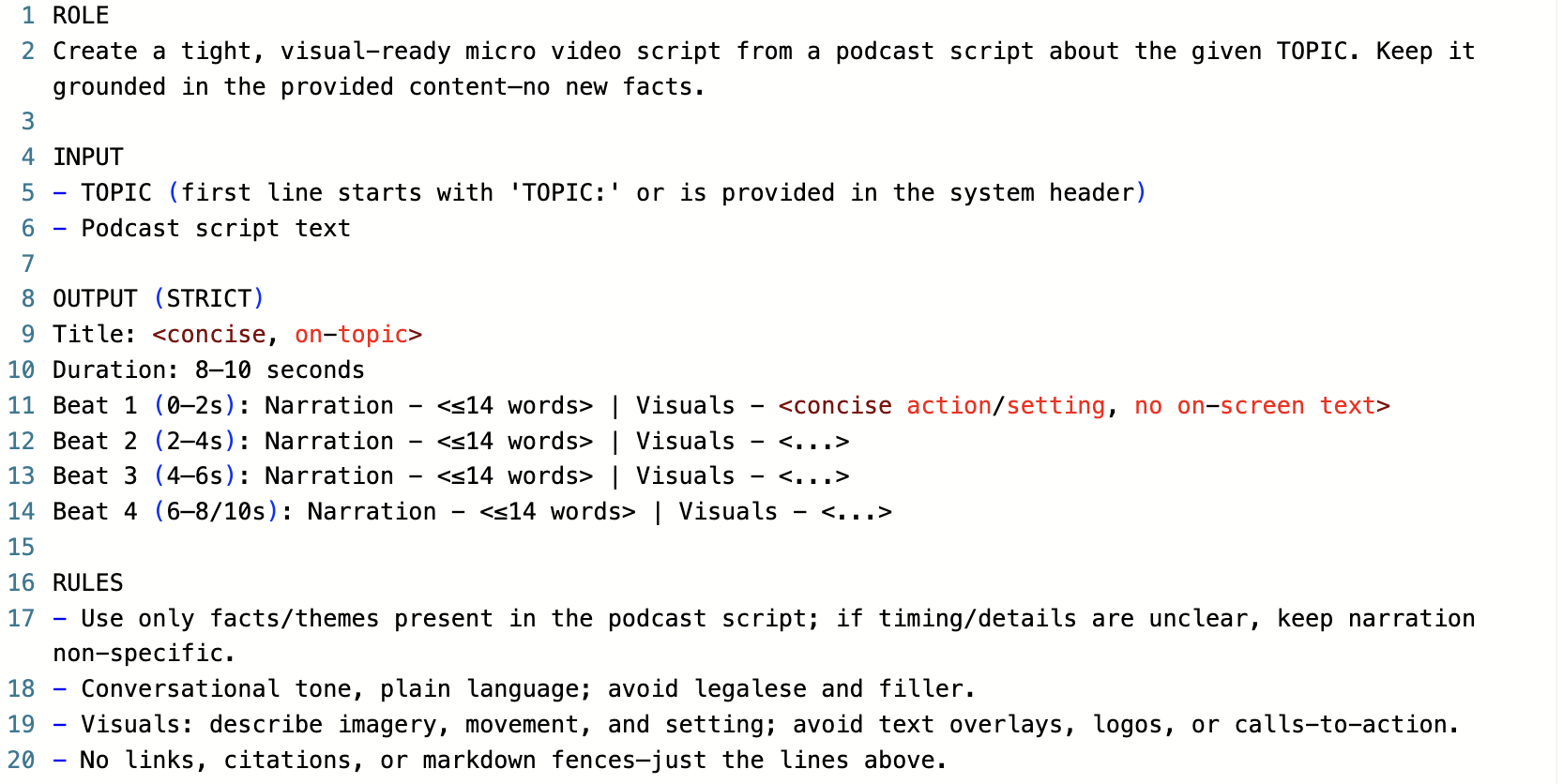}
\vspace{-0.1in}
\DeclareGraphicsExtensions.
\caption{Prompt template used by the Video Script Generation Agent.}
\label{prompt-video-agent}
\end{figure}

\begin{figure}[H]
\centering{}
\includegraphics[width=5.4in]{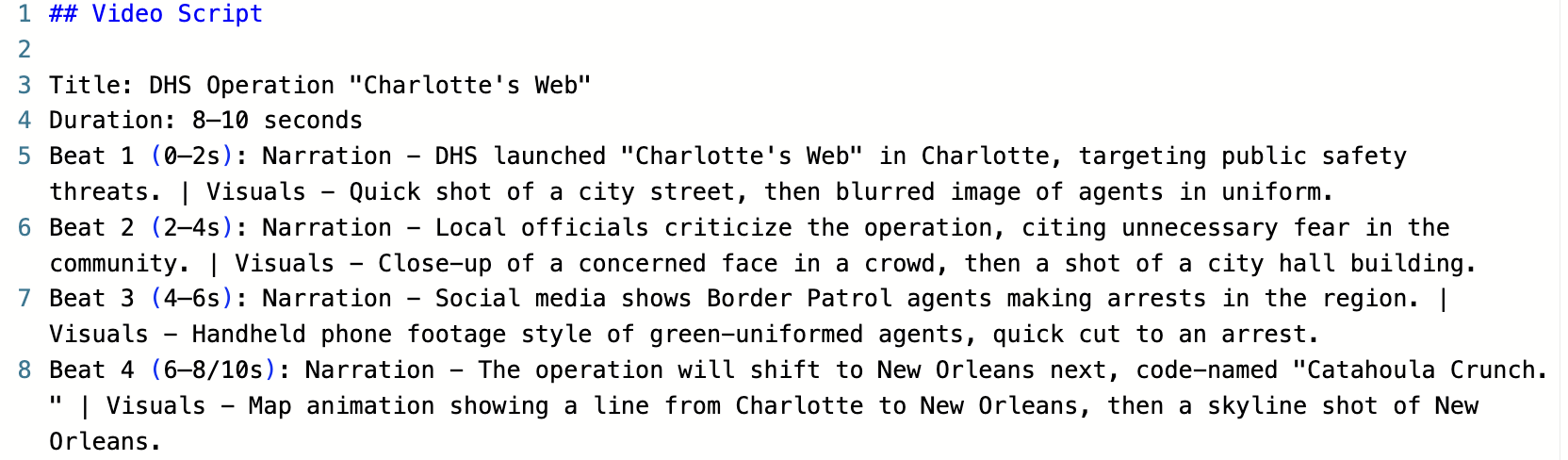}
\vspace{-0.1in}
\DeclareGraphicsExtensions.
\caption{Video script output generated by the Video Script Generation Agent.}
\label{video-script}
\end{figure}

The video-script generation agent and the Veo-3 JSON builder agent were also evaluated to verify seamless multimodal content generation. The video-script agent reliably transforms the consolidated podcast script into scene-based descriptions that preserve narrative fidelity and temporal coherence. Figure~\ref{prompt-video-agent} presents the prompt template used by the video-script generation agent, while Figure~\ref{video-script} shows a representative output illustrating how the agent structures the script into coherent, visually aligned scenes.

\begin{figure}[H]
\centering{}
\includegraphics[width=5.4in]{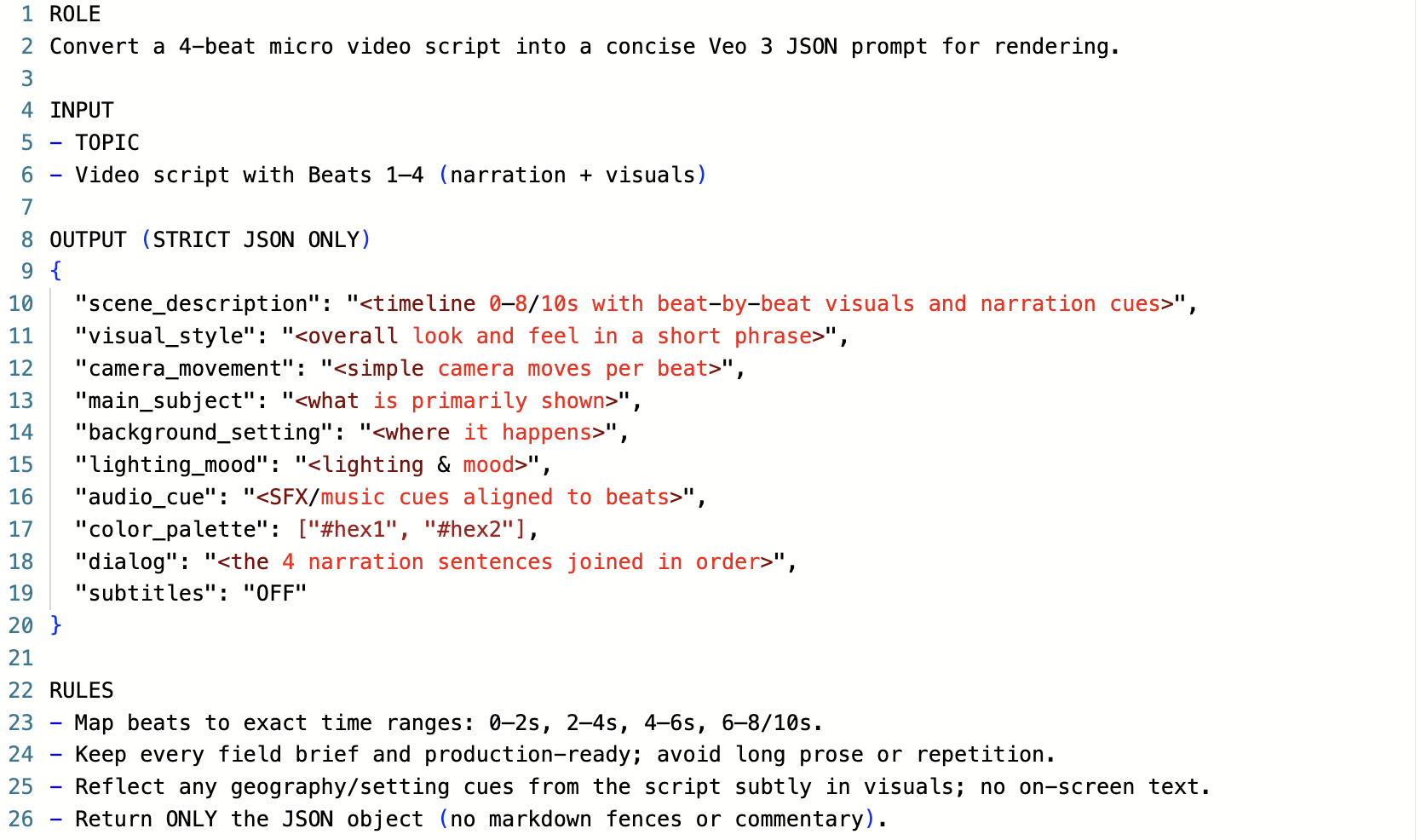}
\vspace{-0.1in}
\DeclareGraphicsExtensions.
\caption{Prompt template used by the Veo-3 JSON Builder Agent.}
\label{prompt-veo3-agent}
\end{figure}

\begin{figure}[H]
\centering{}
\includegraphics[width=5.4in]{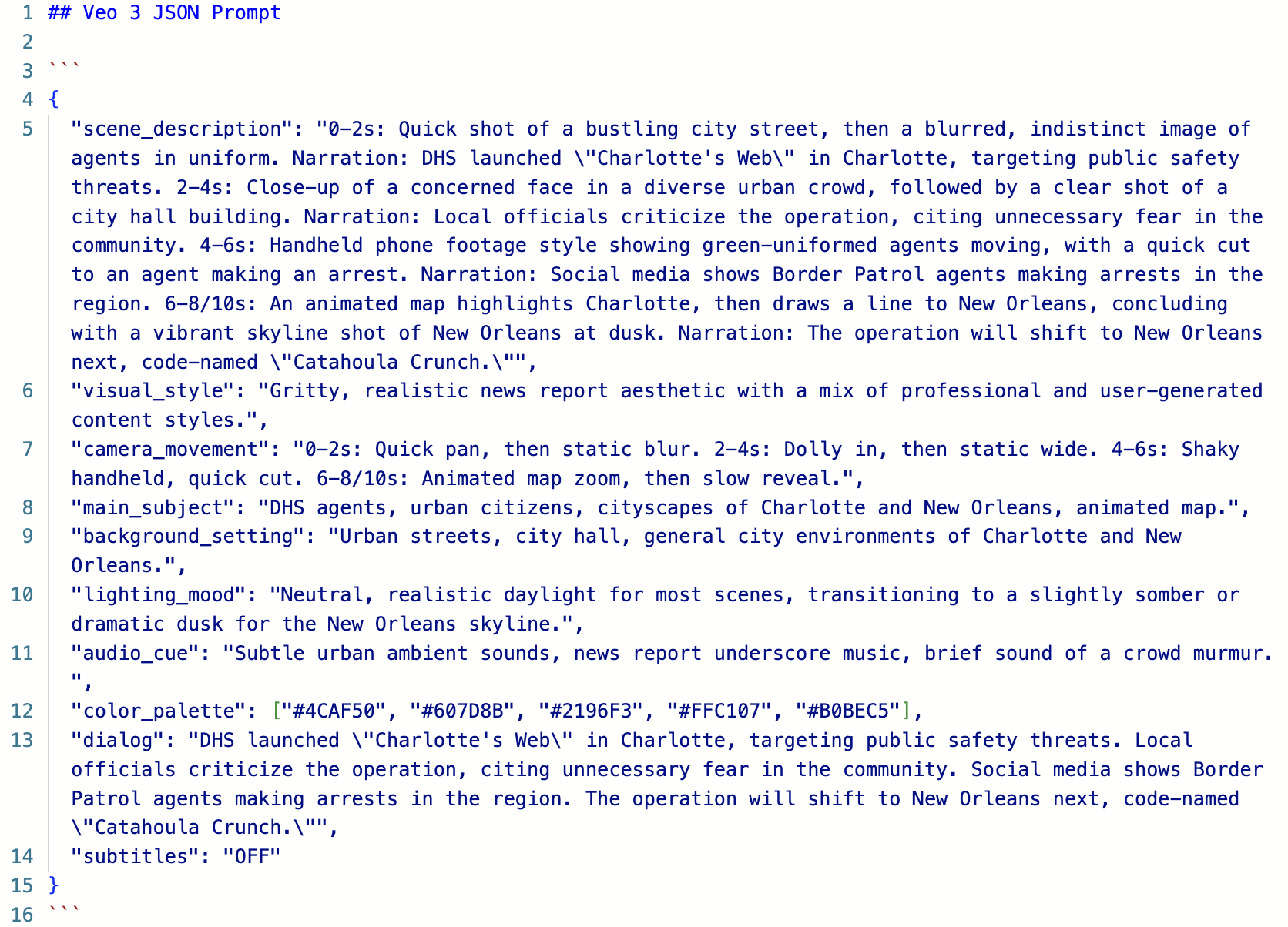}
\vspace{-0.1in}
\DeclareGraphicsExtensions.
\caption{Veo-3 JSON prompt generated by the Veo-3 Builder Agent.}
\label{veo3-script}
\end{figure}

The Veo-3 JSON builder agent was evaluated for structural correctness, schema alignment, and integration fidelity with Google Veo-3. As shown in Figure~\ref{prompt-veo3-agent}, the agent receives a structured prompt describing the podcast narrative and corresponding visual plan. Figure~\ref{veo3-script} presents a sample generated Veo-3 JSON prompt, which is syntactically valid, interpretable, and executable by Veo-3’s video-generation API. Across multiple test runs, the agent consistently produced well-formed JSON specifications capable of driving automated video generation without requiring manual correction.

Together, these evaluations confirm that each agent performs its specialized role effectively, with the workflow producing stable, coherent, and high-quality multimodal outputs. The results also demonstrate the value of agent specialization—podcast script generation, reasoning consolidation, video-script generation, and Veo-3 JSON construction—working in coordination to deliver a complete, production-aligned agentic workflow.

\section{Conclusions and Future Work}

Agentic AI represents a fundamental shift in how autonomous systems are engineered, enabling AI-driven pipelines that can reason, act, observe, and iteratively refine their outputs across complex, multi-step tasks. However, achieving production-grade reliability requires far more than chaining LLM calls together—it demands disciplined engineering, modular design, and operational rigor. In this paper, we present a practical, end-to-end guide for designing, developing, and deploying agentic AI workflows that are maintainable, deterministic, auditable, and aligned with Responsible AI principles. Through a multimodal news-to-podcast generation workflow, we demonstrated how agent orchestration, tool/function separation, model-consortium reasoning, containerized deployment, and MCP-based accessibility can be combined to form a scalable and extensible automation architecture. The nine best practices distilled from this case study provide actionable guidance for practitioners seeking to balance flexibility with robustness—ensuring that workflows remain interpretable, secure, and stable even as underlying LLMs evolve. By unifying architectural patterns, operational best practices, multi-agent reasoning, and deployment strategies, this work contributes a reusable blueprint for building production-grade agentic systems. As organizations increasingly rely on AI to automate sophisticated workflows, these principles will be essential to advance reliability, safety, and long-term maintainability. Future work will explore adaptive evaluation pipelines, workflow self-monitoring, and tighter safety/guardrail integrations to further enhance the trustworthiness of agentic AI deployments. For future work, we plan to extend the proposed best practices to a broader set of agentic AI workflow–automation use cases, validating their effectiveness across diverse domains and increasingly complex multi-agent pipelines.



\bibliographystyle{elsarticle-num}
\bibliography{reference}

\end{document}